\documentclass[final,5p,times,twocolumn]{elsarticle}
\usepackage{array}
\usepackage[utf8x]{inputenc}
\usepackage[T1]{fontenc}
\usepackage[normalem]{ulem}
\usepackage{booktabs}
\usepackage{amsmath,amssymb,amsthm}
\usepackage{bbm}
\usepackage{lineno,hyperref}\hypersetup{colorlinks=true,hypertexnames=false}
\modulolinenumbers[5]
\usepackage{graphicx}
\graphicspath{{./Fig/}}

\usepackage[usenames,dvipsnames]{xcolor}\definecolor{bg}{rgb}{0.95,0.95,0.95}
\usepackage[english]{babel}
\usepackage{listings}
\usepackage{multirow}
\usepackage{tabularx}

\usepackage[linesnumbered,ruled]{algorithm2e}

\usepackage{afterpage}

\usepackage{pgfplotstable}
\usepgfplotslibrary{fillbetween}
\usetikzlibrary{positioning}
\usetikzlibrary{fit}
\usepgfplotslibrary{colorbrewer}
\pgfplotsset{cycle list/Dark2}
\pgfplotsset{compat=newest}
\usepackage{colortbl}

\newcommand{\rev}[1]{\textcolor{black}{#1}}
\newcommand{\revBis}[1]{\textcolor{black}{#1}}
\newcommand{\revTer}[1]{\textcolor{black}{#1}}
\DeclareMathOperator*{\argmin}{argmin}
\DeclareMathOperator*{\argmax}{argmax}
\newcommand\norm[1]{\left\lVert#1\right\rVert}
\newcommand\prox[2]{\text{prox}_{#1}^{#2}}
\newcommand\costR{\mathcal{J}_{\mathrm{r}}}
\newcommand\costC{\mathcal{J}_{\mathrm{c}}}
\newcommand\costG{\mathcal{J}_{\mathrm{g}}}
\newcommand\sigm{\textrm{sigm}}

\begin{document}

\begin{frontmatter}

\title{Matrix Cofactorization for Joint Representation Learning and Supervised Classification -- Application to Hyperspectral Image Analysis\tnoteref{fund}}

\author[n7]{Adrien Lagrange\corref{cor1}}
\ead{adrien.lagrange@enseeiht.fr}
\author[cesbio]{Mathieu Fauvel}
\ead{mathieu.fauvel@inra.fr}
\author[cnes]{St\'{e}phane May}
\ead{stephane.may@cnes.fr}
\author[lisbon]{Jos\'{e} Bioucas-Dias}
\ead{bioucas@lx.it.pt}
\author[n7,iuf]{Nicolas Dobigeon}
\ead{nicolas.dobigeon@enseeiht.fr}
\address[n7]{University of Toulouse, IRIT/INP-ENSEEIHT Toulouse, BP 7122, 31071 Toulouse Cedex 7, France}
\address[cesbio]{CESBIO, University of Toulouse, CNES/CNRS/INRA/IRD/UPS, BPI 2801, 31401 Toulouse Cedex 9, France}
\address[cnes]{CNES, DCT/SI/AP, 18 Avenue Edouard Belin, 31400 Toulouse, France}
\address[iuf]{Institut Universitaire de France, France}
\address[lisbon]{Instituto  de Telecomunica\c{c}\~{o}es,  Instituto  Superior  T\'{e}cnico,  Universidade  de  Lisboa,  1049-001  Lisbon,  Portugal}
\cortext[cor1]{Corresponding author}
\tnotetext[fund]{Part of this work has been supported by Centre National d'\'{E}tudes Spatiales (CNES), Occitanie Region, EU FP7 through the ERANETMED JC-WATER program (project ANR-15-NMED-0002-02 MapInvPlnt), by the ANR-3IA Artificial and Natural Intelligence Toulouse Institute (ANITI) and by the European Research Council (ERC) under the European Union's Horizon 2020 research and innovation programme under grant agreement No 681839 (project FACTORY)}

\begin{abstract}
Supervised classification and representation learning are two widely used classes of methods to analyze multivariate images. Although complementary, these  methods have been scarcely considered jointly in a hierarchical modeling. In this paper, a method coupling these two approaches is designed using a matrix cofactorization formulation. Each task is modeled as a factorization matrix problem and a term relating both coding matrices is then introduced to drive an appropriate coupling. The link can be interpreted as a clustering operation over the low-dimensional representation vectors. The attribution vectors of the clustering are then used as features vectors for the classification task, i.e., the coding vectors of the corresponding factorization problem. A proximal gradient descent algorithm, ensuring convergence to a critical point of the objective function, is then derived to solve the resulting non-convex non-smooth optimization problem. An evaluation of the proposed method is finally conducted both on synthetic and real data in the specific context of hyperspectral image interpretation, unifying two standard analysis techniques, namely unmixing and classification.
\end{abstract}

\begin{keyword}
Image interpretation \sep supervised learning \sep representation learning \sep hyperspectral images \sep non-convex optimization \sep matrix cofactorization
\end{keyword}

\end{frontmatter}


\section{Introduction}
\label{sec:intro}

Numerous frameworks have been developed to efficiently analyze the increasing amount of remote sensing images~\cite{Benediktsson2015,Moser2018}. Among those methods, supervised classification has received considerable attention leading to the development of current state-of-the-art classification methods based on advanced statistical tools, such as convolutional neural networks~\cite{Chen2016,Yuan2019,Lu2019}, kernel methods~\cite{Camps-Valls2009}, random forest~\cite{Belgiu2016} or Bayesian models~\cite{Chang2019}. In the context of remote sensing image classification, these methods aim at retrieving the class of each pixel of the image given a specific class nomenclature. Within a supervised framework, a set of pixels is assumed to be annotated by an expert and subsequently used as examples through a learning process. Thanks to extensive research efforts of the community, classification methods have become very efficient. Nevertheless, they still face some challenging issues, such as the high dimension of the data, often coupled with the lack of training data~\cite{Hughes1968}. Handling multi-modal and/or composite classes with intrinsic intra-variability is also a recurrent issue~\cite{Qi2019}: for instance, a class referred to as \emph{building} can gather very dissimilar samples when metallic and tiled roofs are present in a scene. Besides, the resulting classification remains a high-level interpretation of the scene since it only gives a single class to summarize all information in a given pixel.

Hence, more recent works have emerged in order to provide a richer interpretation~\cite{Bioucas-Dias2012,Eches2011}. In particular, representation learning methods assume that the data results from the composition of a reduced number of elementary patterns. More precisely, the observed measurements can be approximated by mixtures of dictionary elements able to simultaneously capture the variability and redundancy in the dataset. 
Representation learning can be tackled from different perspectives, in particular known as dictionary learning~\cite{Aharon2006}, source separation~\cite{Zibulevsky2001}, compressive sensing~\cite{Porta2019}, factor analysis~\cite{Cavalcanti2018}, matrix factorization~\cite{Koren2009} or subspace learning~\cite{Elhamifar2013}. Various models have been proposed to learn a dedicated representation relevant to the field of interest, differing by specific assumptions and/or constraints. Most of them attempt to identify a dictionary and a mixture function by minimizing a reconstruction error measuring the discrepancy between the chosen model and the dataset. For instance, non-negative matrix factorization (NMF) aims at recovering a linear mixture of non-negative elements with non-negative activation coefficients leading to additive part-based decompositions of the observations~\cite{Lee1999,Donoho2004}. Contrary to a classification task, representation learning methods have generally the great advantage of being unsupervised. However, for particular purposes, they can be specialized to learn a representation suited for a particular task, e.g. classification or regression~\cite{Mairal2012}. \rev{Thus, representation learning provides a rich yet compact description of the data whereas supervised classification offers a univocal interpretation based on prior knowledge from experts.}

The idea of combining the representation learning and classification tasks has already been considered, mostly to use the representation learning method as a dimensionality reduction step prior to the classification~\cite{Zheng2017}, where the low-dimensional representation is used as input features. Nonetheless, some works introduce the idea of performing the two tasks simultaneously~\cite{Zhang2018a}. For example, the discriminative K-SVD algorithm associates a linear mixture model to a linear classifier ~\cite{Zhang2010}. At the end, the method tries to learn a dictionary well-fitted for the classification task, i.e., the learned representation minimizes the reconstruction error but also ensures a good separability of the classes. More intertwined frameworks can be also considered, as the one proposed in~\cite{Jiang2011} where elements of the dictionary are class-specific. Joint representation learning and classification can be cast as a cofactorization problem. Both tasks are interpreted as individual factorization problems and constraints between the dictionaries and coding matrices associated with the two problems can then be imposed. These cofactorization-based models have proven to be highly efficient in many application fields, e.g. for text mining~\cite{Wang2011}, music source separation~\cite{Yoo2010}, or image analysis~\cite{Yokoya2012,Akhtar2018}.

However, most of the available methods tend to focus on classification results and generally oppose reconstruction accuracy and discriminative abilities of the models instead of designing a unifying hierarchical structure. Capitalizing on recent advances and a first attempt in \cite{Lagrange2018} in a Bayesian setting, this paper proposes a particular cofactorization method, with a dedicated application to multivariate image analysis. The representation learning and classification tasks are related through the coding matrices of the two factorization problems. A clustering is performed on the low-dimensional representation and the clustering attribution vectors are used as coding vectors for the classification. This novel coupling approach produces a coherent and fully-interpretable hierarchical model. To solve the resulting non-convex non-smooth optimization problem, a proximal alternating linearized minimization (PALM) algorithm is derived, yielding guarantees of convergence to a critical point of the objective function \cite{Bolte2014}.

\rev{
The main contributions reported in this paper can be summarized as follows. A generic framework is proposed to demonstrate that two ubiquitous image analysis methods, namely supervised classification and representation learning, can be unified into a unique joint cofactorization problem. This framework is instanced for one particular application in the context of hyperspectral image analysis where supervised classification and spectral unmixing are performed jointly. The proposed method offers a comprehensive and meaningful analysis of the image as well as competitive quantitative results for the two considered tasks.
}

This paper is organized as follows. Section~\ref{sec:pb-stat} defines the two factorization problems used to perform representation learning and classification and further discusses the joint cofactorization problem. It also details the optimization scheme developed to solve the resulting non-convex minimization problem. To illustrate the generic framework introduced in the previous section, an application to hyperspectral image analysis is conducted in Section~\ref{sec:hs-app} through the dual scope of spectral unmixing and classification. Performance of the proposed framework is illustrated thanks to experiments conducted on synthetic and real data in Section \ref{sec:exp}. Finally, Section~\ref{sec:ccl} concludes the paper and presents some research perspectives to this work.

\section{Proposed generic framework}
\label{sec:pb-stat}

The representation learning and classification tasks are generically defined as factorization matrix problems in Sections~\ref{sec:repres-pb} and~\ref{sec:classif-pb}. To derive a unified cofactorization formulation, a third step consists in drawing the link between these two independent problems. In this work, this coupling is ensured by imposing a consistent structure between the two coding matrices corresponding to the low-dimensional representation and the feature matrices, respectively. As detailed in Section~\ref{sec:clust-pb}, it is expressed as a clustering task where the parameters describing the attribution to the clusters are the feature vectors, i.e. the coding matrix resulting from the classification task. Particular instances of these three tasks will be detailed in Section \ref{sec:hs-app} for an application to multiband image analysis.

\subsection{Representation learning}
\label{sec:repres-pb}
The fundamental assumption in representation learning is that the $P$ considered $L$-dimensional samples, gathered in matrix $\mathbf{Y} \in \mathbb{R}^{L\times P}$, belong to a $R$-dimensional subspace such that $R\ll L$. The aim is then to recover this manifold, where samples can be expressed as combinations of elementary vectors, herein the column of the matrix $\mathbf{W} \in \mathbb{R}^{L\times R}$ sometimes referred to as dictionary. These samples can be subsequently represented thanks to the so-called coding matrix $\mathbf{H} \in \mathbb{R}^{R\times P}$. Formally, identifying the dictionary and the coding matrices can be generally expressed as a minimization problem\vspace{-0.15cm}
\begin{align}
 \label{eq:repres-pb-gen}
  \min_{\mathbf{W},\mathbf{H}} \costR(\mathbf{Y} | \psi(\mathbf{W}, \mathbf{H})) &+ \lambda_{w} \mathcal{R}_{w}(\mathbf{W})+ \imath_{\mathbb{W}}(\mathbf{W}) \nonumber \\
  & + \lambda_h \mathcal{R}_{h}(\mathbf{H}) + \imath_{\mathbb{H}}(\mathbf{H})\vspace{-0.15cm}
\end{align}
where $\psi(\cdot)$ is a mixture function (e.g., linear or bilinear operator), $\costR(\cdot)$ is an appropriate cost function, for example derived from a $\beta$-divergence~\cite{Fevotte2011}, $\mathcal{R}_{\cdot}(\cdot)$ denote penalizations weighted by the parameter $\lambda_{\cdot}$ and $\imath_{\cdot}(\cdot)$ is the indicator functions defined here on the respective sets $\mathbb{W}\subset \mathbb{R}^{L\times R}$ and $\mathbb{H}\subset \mathbb{R}^{R\times P}$ imposing some constraints on the dictionary and coding matrices.

In the case of a linear embedding adopted in this work, the mixture function writes\vspace{-0.15cm}
\begin{equation}
\label{eq:psi}
\psi(\mathbf{W},\mathbf{H}) = \mathbf{W}\mathbf{H}.\vspace{-0.15cm}
\end{equation}
In this context, the problem \eqref{eq:repres-pb-gen} can be cast as a factor analysis driven by the cost function $\costR(\cdot)$. Depending on the applicative field, typical data-fitting measures include the Itakura-Saito, the Euclidean and the Kullback-Leibler divergences~\cite{Fevotte2011}. Assuming a low-rank model (i.e., $R \leq L$), specific choices for the sets $\mathbb{H}$ and $\mathbb{W}$ lead to various standard factor models. For instance, when $\mathbb{W}$ is chosen as the Stiefel manifold, the solution of \eqref{eq:repres-pb-gen} is given by a principal component analysis (PCA)~\cite{Jolliffe2011}. When $\mathbb{W}$ and $\mathbb{H}$ impose nonnegativity of the dictionary and coding matrix elements, the problem is known as nonnegative matrix factorization~\cite{Lee1999,Paatero1994}.

Within a supervised context, the dictionary  $\mathbf{W}$ can be chosen thanks to a end-user expertise or estimated beforehand. Without loss of generality but for the sake of conciseness, the framework described in this paper assumes that this dictionary is known, possibly overcomplete as proposed in the experimental illustration described in Section~\ref{sec:exp}.
In this case, as in many applications, it makes sense to look for a sparse representation of the signal of interest to retrieve its most achievable compact representation~\cite{Mairal2012,Bruckstein2008}. Following this strategy, we propose to consider an $\ell_1$-norm sparsity penalization on the coding vectors, leading to representation learning task defined by\vspace{-0.15cm}
\begin{equation}
\label{eq:repres-pb}
  \min_{\mathbf{H}}  \costR(\mathbf{Y} | \mathbf{W}\mathbf{H}) + \lambda_{h} \norm{\mathbf{H}}_1 + \imath_{\mathbb{H}}(\mathbf{H})\vspace{-0.15cm}
\end{equation}
where $\norm{\mathbf{H}}_1 = \sum_{p=1}^P \norm{\mathbf{h}_p}_1$ with $\mathbf{h}_p$ denoting the $p$th column of $\mathbf{H}$.

\subsection{Supervised classification}
\label{sec:classif-pb}


To clearly define the classification task, let first introduce some key notations. The index subset of samples with an available groundtruth is denoted as $\mathcal{L}$ while the index subset of unlabeled samples is $\mathcal{U}$ such that $\mathcal{L} \cap \mathcal{U} = \emptyset$ and $\mathcal{L} \cup \mathcal{U} = \mathcal{P}$ with $\mathcal{P} \triangleq \left\{1,\ldots,P\right\}$. Classifying the unlabeled samples consists in assigning each of them to one of the $C$ classes. This can be reformulated as the estimation of a $C\times P$ matrix $\mathbf{C}$ whose columns correspond to unknown $C$-dimensional attribution vectors $\mathbf{c}_p=\left[c_{1,p},\ldots,c_{C,p}\right]^T$. Each vector is made of $0$ except for $c_{i,p}=1$ when the $p$th sample is assigned the $i$th class.

Numerous classification rules have been proposed in the literature~\cite{Hastie2009}. Most of them rely on a $K\times P$ matrix $\mathbf{Z}=\left[\mathbf{z}_1,\ldots,\mathbf{z}_P\right]$ of features $\mathbf{z}_p$ ($p\in\mathcal{P}$) associated with each sample and derived from the raw data. Within a supervised framework, the attribution matrix $\mathbf{C}_{\mathcal{L}}$ and feature matrix $\mathbf{Z}_{\mathcal{L}}$ of the labeled data are exploited during the learning step, where $\cdot_{\mathcal{L}}$ denotes the corresponding submatrix whose columns are indexed by ${\mathcal{L}}$. For a wide range of classifiers, deriving a classification rule can be achieved by solving the optimization problem\vspace{-0.15cm}
\begin{equation}
\label{eq:classif-pb0}
  \min_{\mathbf{Q}} \costC(\mathbf{C}_\mathcal{L}| \phi(\mathbf{Q},\mathbf{Z}_\mathcal{L})) + \lambda_{q} \mathcal{R}_{q}(\mathbf{Q})\vspace{-0.15cm}
\end{equation}
where  $\mathbf{Q} \in\mathbb{R}^{C\times K}$ is the set of classifier parameters to be inferred, $\mathcal{R}_{q}(\cdot)$ refer to regularizations imposed on $\mathbf{Q}$ and $\costC$ is a cost function  measuring the quality of the classification such as the quadratic loss~\cite{Zhang2010} or cross-entropy~\cite{Kline2005}. Moreover, in \eqref{eq:classif-pb0}, $\phi(\mathbf{Q},\cdot)$ defines a element-wise nonlinear mapping between the features and the class attribution vectors parametrized by $\mathbf{Q}$, e.g., derived from a sigmoid or a softmax operators. In this work, the classifier is assumed to be linear, which leads to a vector-wise post-nonlinear mapping\vspace{-0.15cm}
\begin{equation}
\label{eq:phi}
\phi(\mathbf{Q},\mathbf{Z}_\mathcal{L}) = \phi(\mathbf{Q}\mathbf{Z}_\mathcal{L})\vspace{-0.15cm}
\end{equation}
with\vspace{-0.15cm}
\begin{equation}
\phi(\mathbf{X}) = \left[\phi(\mathbf{x}_1),\ldots,\phi(\mathbf{x}_p)\right].\vspace{-0.15cm}
\end{equation}

Once the classifier parameters have been estimated by solving~\eqref{eq:classif-pb0}, the unknown attribution vectors $\mathbf{C}_{\mathcal{U}}$ can be subsequently inferred during the testing step by applying the nonlinear transformation to the corresponding predicted features $\hat{\mathbf{Z}}_{\mathcal{U}}$ associated with the unlabeled samples. The obtained outputs are relaxed attribution vectors $\hat{\mathbf{c}}_p = \phi(\mathbf{Q}\hat{\mathbf{z}}_p)$ ($p\in \mathcal{U}$) and the most probable predicted sample class can be computed as $\argmax_i c_{i,p}$.

Under the proposed formulation of the classification task, the learning and testing steps can be conducted simultaneously, a framework usually referred to as semi-supervised, with the beneficial opportunity to introduce additional regularizations and/or constraints on the submatrix of unknown attribution vectors $\mathbf{C}_\mathcal{U}$.  The initial problem \eqref{eq:classif-pb0} is thus extended to the following one\vspace{-0.15cm}
\begin{equation}
\label{eq:classif-pb1}
  \min_{\mathbf{Q},\mathbf{C}_\mathcal{U}} \costC(\mathbf{C}| \phi(\mathbf{Q}\mathbf{Z})) + \lambda_{q}  \mathcal{R}_{q}(\mathbf{Q}) + \lambda_{c} \mathcal{R}_{c}(\mathbf{C})
  + \imath_{\mathbb{C}}(\mathbf{C}_\mathcal{U})\vspace{-0.15cm}
\end{equation}
where $\mathbf{C} = [\mathbf{C}_{\mathcal{L}}\ \mathbf{C}_{\mathcal{U}}]$ and $\mathbb{C} \subset \mathbb{R}^{C\times |\mathcal{U}|}$ denotes a feasible set for the attribution matrix $\mathbf{C}_\mathcal{U}$. \rev{As discussed above, the cost function $\costC(\mathbf{C}|  \hat{\mathbf{C}} )$ measures the actual classification loss, i.e., the discrepancy between the attribution vector $\mathbf{C}$ of the training set and the attribution vectors $\mathbf{\hat{C}}$ predicted by the classifier. Two particular cases fitting this generic model are provided in Section~\ref{sec:classif_quadratic} and~\ref{sec:classif_entropy}. The attribution vectors are defined as $\hat{\mathbf{C}} = \phi(\mathbf{Q}\hat{\mathbf{Z}})$ where $\phi(\cdot)$ is a nonlinear function applied to the output of a linear classifier. The regularization term $\mathcal{R}_{q}(\mathbf{Q})$ penalizes over the parameters of the classifiers. A typical example is a quadratic penalization which aims at avoiding overfitting, as conventionally done when optimizing neural networks and generally referred to as \emph{weight decay}~\cite{Goodfellow2016}. Finally, the regularization term $\mathcal{R}_{c}(\mathbf{C})$ penalizes over the attribution matrix. Typical examples include spatial regularizations such as total variation (TV) when dealing with image classification. The indicator function $\imath_{\mathbb{C}}(\mathbf{C}_\mathcal{U})$ enforces sum-to-one and non-negativity constraints such that each attribution vector $\mathbf{c}_p$ ($p\in \mathcal{U}$) can then be interpreted as a probability vector of belonging to each class. In such a case, the feasible set is chosen as $\mathbb{C} = \mathbb{S}_C^{|\mathcal{U}|}$ where\vspace{-0.15cm}
\begin{equation}
\label{eq:simplexC}
   \mathbb{S}_C \triangleq \left\{\mathbf{u} \in \mathbb{R}^C \big| \forall k,\  u_k \geq 0 \ \text{and} \ \sum_{k=1}^C u_k=1\right\}.\vspace{-0.15cm}
\end{equation}
}

\subsection{Coupling representation learning and classification}
\label{sec:clust-pb}
\begin{figure}
  \centering
  \begin{tabular}{m{2.6cm}|@{\hspace{0.1cm}}c@{\hspace{0.1cm}}|m{3cm}}
    \textsc{Representation}

    \textsc{learning} &
    \textsc{Clustering} &
    \textsc{Classification} \\
    \begin{tikzpicture}[auto, semithick]
      \node[rectangle,draw,minimum width=1.3cm,minimum height=1.5cm] (A) {$\textcolor{blue}{\mathbf{Y}}$};
      \node[rectangle,draw,above=0.25 of A,minimum width=1.3cm,minimum height=1cm] (B) {$\textcolor{olive}{\mathbf{H}}$};
      \node[rectangle,draw,dotted,left=0.25 of A,minimum width=0.5cm,minimum height=1.5cm] (C) {$\mathbf{W}$};

      \node[below=0.25 of A] (L1) {\footnotesize Image};
      \node[above=0.25 of B] (L2) {\footnotesize Codes};
      \node[below=0.25 of C] (L3) {\footnotesize Dict.};
    \end{tikzpicture} &
    $\min_\mathbf{Z} \costG(\textcolor{olive}{\mathbf{H}}, \textcolor{olive}{\mathbf{Z}} ; \boldsymbol{\theta})$ &
    \begin{tikzpicture}[auto, semithick]
      \node[rectangle,draw,minimum width=1.4cm,minimum height=0.75cm] (A) {$\begin{array}{@{}c@{\hspace{0.1cm}}|@{\hspace{0.05cm}}c@{}} \textcolor{blue}{\mathbf{C}_\mathcal{L}} & \mathbf{C}_\mathcal{U} \\ \end{array}$};
      \node[rectangle,draw,above=0.25 of A,minimum width=1.4cm,minimum height=1.25cm] (B) {$\textcolor{olive}{\mathbf{Z}}$};
      \node[rectangle,draw,left=0.25 of A,minimum width=0.5cm,minimum height=0.75cm] (C) {$\mathbf{Q}$};

      \node[below right=0.25 and -1.6 of A] (L1) {\footnotesize Classification};
      \node[above=0.25 of B] (L2) {\footnotesize Features};
      \node[below left=0.25 and -0.7 of C] (L3) {\footnotesize Classifier};
    \end{tikzpicture} \\
  \end{tabular}
  \caption{Structure of the cofactorization model. Variables in \textcolor{blue}{\emph{blue}} stand for observations or available external data. Variables in \textcolor{olive}{\emph{olive green}} are linked through the clustering task here formulated as an optimization problem. The variable in a dotted box is assumed to be known or estimated beforehand in this work. \label{fig:model}}
\end{figure}
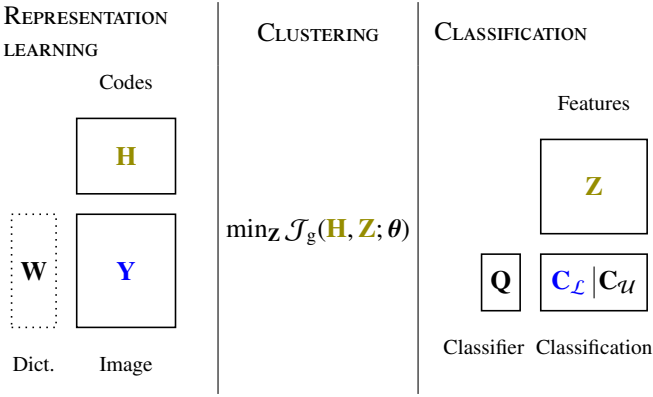

Up to this point, the representation learning and supervised classification tasks have been formulated as two independent matrix factorization problems given by \eqref{eq:repres-pb} and \eqref{eq:classif-pb1}, respectively. This work proposes to join them by drawing an implicit relation between two factors involved in these two problems. Inspired by hierarchical Bayesian models such as the one proposed in~\cite{Lagrange2018}, both problems are coupled through the activation matrices $\mathbf{H}$ and $\mathbf{Z}$, as illustrated in Figure~\ref{fig:model}. More precisely, the coding vectors in $\mathbf{H}$ are clustered such that the feature vectors in $\mathbf{Z}$ are defined as the attribution vectors to the $K$ clusters. Ideally, clustering attribution vectors $\mathbf{z}_p$ are filled with zeros except for $z_{k,p}=1$ when $\mathbf{h}_p$ is associated with the $k$th cluster. Thus, the vectors $\mathbf{z}_p$ ($p\in \mathcal{P}$) are assumed to be defined on the $K$-dimensional probability simplex $\mathbb{S}_K$ similarly defined as \eqref{eq:simplexC} and ensuring non-negativity and sum-to-one constraints. Many clustering algorithms can be expressed as optimization problem such as the well-known k-means algorithm and many of its variants \cite{Condat2017,Pompili2014}. Adopting this formulation, and denoting $\boldsymbol{\theta}$ the set of parameters of the clustering algorithm, the clustering task can be defined as the minimization problem\vspace{-0.15cm}
\begin{equation}
\label{eq:clust-pb}
  \min_{\mathbf{Z},\boldsymbol{\theta}  } \costG(\mathbf{H}, \mathbf{Z} ; \boldsymbol{\theta}) + \lambda_{z} \mathcal{R}_{z}(\mathbf{Z})+ \lambda_{\theta} \mathcal{R}_{\theta}(\boldsymbol{\theta}) + \imath_{\mathbb{S}_K^P}(\mathbf{Z}) + \imath_{\boldsymbol{\Theta}}(\boldsymbol{\theta})\vspace{-0.15cm}
\end{equation}
where $\boldsymbol{\Theta}$ defines a feasible set for the parameters $\boldsymbol{\theta}$.

It is worth noting that introducing this coupling term is one of the major novelty of the proposed approach. When considering task-driven dictionary learning methods, it is usual to intertwine the representation learning and the classification tasks by directly imposing $\mathbf{H}=\mathbf{Z}$~\cite{Zhang2010,Sun2015}. Since these methods generally rely on a linear classifier, one major drawback of such approaches is their unability to deal with non-separable classes in the low-dimensional representation space. In such cases, the underlying model cannot be discriminative and descriptive simultaneously and the resulting tasks become adversarial. When considering the proposed coupling term, the cluster attribution vectors $\mathbf{z}_p$ offer the possibility of linearly separating any group of clusters from the others. As a consequence, the model benefits from more flexibility, with both discriminative and descriptive abilities in a more general sense.

\subsection{Global cofactorization problem}
\label{sec:cofact-pb}

Unifying the representation learning task \eqref{eq:repres-pb} and the classification task \eqref{eq:classif-pb1} through the clustering task \eqref{eq:clust-pb} leads to the following joint cofactorization problem
\begin{align}
\label{eq:glob-pb}
  &\min_{\substack{ \mathbf{H},\mathbf{Q},\mathbf{C}_{\mathcal{U}},\\ \mathbf{Z},\boldsymbol{\theta}}  }
  \lambda_0 \costR(\mathbf{Y} | \mathbf{W}\mathbf{H}) + \lambda_{h} \norm{\mathbf{H}}_1 \nonumber \\
  & + \lambda_1  \costC(\mathbf{C}| \phi(\mathbf{Q}\mathbf{Z})) +  \lambda_{q}  \mathcal{R}_{q}(\mathbf{Q}) + \lambda_{c} \mathcal{R}_{c}(\mathbf{C})  \nonumber \\
  & + \lambda_2 \costG(\mathbf{H}, \mathbf{Z} ; \boldsymbol{\theta}) + \lambda_{z} \mathcal{R}_{z}(\mathbf{Z})+ \lambda_{\theta} \mathcal{R}_{\theta}(\boldsymbol{\theta})  \nonumber\\
  & + \imath_{\mathbb{H}}(\mathbf{H}) + \imath_{\mathbb{S}_K^{|\mathcal{U}|}}(\mathbf{C}_{\mathcal{U}}) + \imath_{\mathbb{S}_K^P}(\mathbf{Z}) + \imath_{\boldsymbol{\Theta}}(\boldsymbol{\theta})
\end{align}
where $\lambda_0$, $\lambda_1$ and $\lambda_2$ control the respective contribution of each task data-fitting term. All notations and parameter dimensions are summarized in Table \ref{table:notations}. A generic algorithmic scheme solving the problem \eqref{eq:glob-pb} is proposed in the next section.

\begin{table}[!ht]
\centering
\caption{Overview of notations. \label{table:notations}}
\begin{tabular}{r@{} l |c}
  \hline
    &  & parameter\\
    \hline
  $P$\            & $\in \mathbb{R}$             & number of observations\\
  $L$\            & $\in \mathbb{R}$             & dimension of observations\\
  $C$\            & $\in \mathbb{R}$             & number of classes\\
  $K$\            & $\in \mathbb{R}$             & number of features/clusters\\
  $\mathcal{P}$\  & $=\left\{1,\ldots,P\right\}$                         & index set of observations\\
  $\mathcal{L}$\  & $\subset \mathcal{P}$                         & index set of labeled samples\\
  $\mathcal{L}_i$\  & $\subset \mathcal{L}$                         & index set of labeled samples in the $i$th class\\
  $\mathcal{U}$\  & $= \mathcal{P}\backslash \mathcal{L}$                         & index set of unlabeled samples\\
  $\mathbf{Y}$\  & $\in \mathbb{R}^{L\times P}$ & observations \\
  $\mathbf{W}$\  & $\in \mathbb{R}^{L\times R}$ & dictionary \\
  $\mathbf{H}$\  & $\in \mathbb{R}^{R\times P}$ & coding matrix \\
  $\mathbf{Q}$\  & $\in \mathbb{C}^{C\times P}$ & classifier parameters\\
  $\mathbf{C}_{\mathcal{L}}$\  & $\in \mathbb{R}^{C\times |\mathcal{L}|}$ & attribution matrix of labeled data\\
  $\mathbf{C}_{\mathcal{U}}$\  & $\in \mathbb{R}^{C\times |\mathcal{U}|}$ & attribution matrix of unlabeled data\\
  $\mathbf{C}$\  & $= \left[\mathbf{C}_{\mathcal{L}} \ \mathbf{C}_{\mathcal{U}}\right]$ & class attribution matrix\\

  $\mathbf{Z}$\  & $\in \mathbb{R}^{K\times P}$ & cluster attribution matrix\\
  $\boldsymbol{\theta}$\  & $\in \boldsymbol{\Theta}$ & clustering parameters\\
  \hline
\end{tabular}
\end{table}

\subsection{Optimization scheme}
\label{sec:optim}

The minimization problem defined by \eqref{eq:glob-pb} is not globally convex. To reach a local minimizer, we propose to resort to the proximal alternating linearized minimization (PALM) algorithm introduced in \cite{Bolte2014}. \rev{This algorithm is based on proximal descent steps, which allows non-smooth terms to be handled. Moreover it is guaranteed to converge to a critical point of the objective function even in the case of non-convex problem.} This means that, if the initialization is good enough, it is expected to likely converge to a solution close to the global optimum. To implement PALM, the problem \eqref{eq:glob-pb} is rewritten in the form of an unconstrained problem expressed as a sum of a smooth coupling term $g(\cdot)$ and separable non-smooth terms $f_j(\cdot)$ ($j\in\left\{0,\ldots,4\right\}$) as follows
\begin{align}
\label{eq:palm-pb}
  \min_{\substack{ \mathbf{H},\boldsymbol{\theta},\mathbf{Z},\\\mathbf{Q},\mathbf{C}_{\mathcal{U}}} } & \ f_0(\mathbf{H}) + f_1(\boldsymbol{\theta}) + f_2(\mathbf{Z}) + f_3(\mathbf{C}_{\mathcal{U}}) \nonumber\\
  & + g(\mathbf{H},\boldsymbol{\theta},\mathbf{Z},\mathbf{C}_{\mathcal{U}},\mathbf{Q})\vspace{-0.15cm}
\end{align}
where
\begin{equation}\label{eq:functions_f}
\begin{array}{rl}
f_0(\mathbf{H}) &= \imath_{\mathbb{H}}(\mathbf{H}) + \lambda_{h} \norm{\mathbf{H}}_1 \nonumber\\
  f_1(\boldsymbol{\theta}) &= \imath_{\boldsymbol{\Theta}}(\boldsymbol{\theta}) \nonumber\\
      \end{array}
    \quad
\begin{array}{rl}
  f_2(\mathbf{Z}) &= \imath_{\mathbb{S}_K^P}(\mathbf{Z}) \nonumber\\
  f_3(\mathbf{C}_{\mathcal{U}}) &= \imath_{\mathbb{S}_K^{|\mathcal{U}|}}(\mathbf{C}_{\mathcal{U}}) \nonumber\\
 \end{array}\vspace{-0.15cm}
\end{equation}
and the coupling function is
\begin{align}
  g(\mathbf{H},\boldsymbol{\theta}&,\mathbf{Z},\mathbf{C}_{\mathcal{U}},\mathbf{Q}) = \lambda_0 \costR(\mathbf{Y} | \mathbf{W}\mathbf{H}) \nonumber\\
  &+ \lambda_1 \costC(\mathbf{C}| \phi(\mathbf{Q}\mathbf{Z}))
  +  \lambda_{q}  \mathcal{R}_{q}(\mathbf{Q}) + \lambda_{c} \mathcal{R}_{c}(\mathbf{C}) \nonumber\\
  &+ \lambda_2 \costG(\mathbf{W}, \mathbf{Z} ; \boldsymbol{\theta})
  + \lambda_{z} \mathcal{R}_{z}(\mathbf{Z})+ \lambda_{\theta} \mathcal{R}_{\theta}(\boldsymbol{\theta}).
\end{align}

To ensure the stated guarantees of PALM, all $f_j(\cdot)$ have to be proper, lower semi-continuous function $f_j: \mathbb{R}^{n_j} \rightarrow (-\infty,+\infty]$, which ensures in particular that the associated proximal operator is well-defined. Additionally, sufficient conditions on the coupling function are that $g(\cdot)$ is a $\mathcal{C}^2$ function (i.e., with continuous first and second derivatives) and that its partial gradients are globally Lipschitz. For example, partial gradient $\nabla_\mathbf{H} g(\mathbf{H},\boldsymbol{\theta},\mathbf{Z},\mathbf{C}_{\mathcal{U}},\mathbf{Q})$ should be globally Lipschitz for any fixed $\boldsymbol{\theta}$, $\mathbf{Z}$, $\mathbf{C}_{\mathcal{U}}$, $\mathbf{Q}$, that is
\begin{align}
  &\norm{\nabla_\mathbf{H} g(\mathbf{H}_1,\boldsymbol{\theta},\mathbf{Z},\mathbf{C}_{\mathcal{U}},\mathbf{Q}) - \nabla_\mathbf{H} g(\mathbf{H}_2,\boldsymbol{\theta},\mathbf{Z},\mathbf{C}_{\mathcal{U}},\mathbf{Q})} \leq \nonumber\\
  & \quad L_\mathbf{H}(\boldsymbol{\theta},\mathbf{Z},\mathbf{C}_{\mathcal{U}},\mathbf{Q}) \norm{\mathbf{H}_1 - \mathbf{H}_2}, \quad \forall \mathbf{H}_1,\mathbf{H}_2 \in \mathbb{R}^{R\times P} \label{eq:Lip_constant_def}
\end{align}
where $L_\mathbf{H}(\boldsymbol{\theta},\mathbf{Z},\mathbf{C}_{\mathcal{U}},\mathbf{Q})$, simply denoted $L_\mathbf{H}$ hereafter, is the Lipschitz constant. For sake of conciseness, we refer to \cite{Bolte2014} to get further details.

The main idea of the algorithm is then to update each variable of the problem alternatively using a proximal gradient descent. The overall scheme is summarized in Algorithm~\ref{alg:palm}. For a practical implementation, one needs to compute the partial gradients of $g(\cdot)$ explicitly and their Lipschitz constants to perform a gradient descent step, followed by a proximal mapping associated with the non-smooth terms $f_j(\cdot)$.  The objective function is then monitored at each iteration and the algorithm is stopped when convergence is reached. Note that, when a specific penalization $\mathcal{R}_{\cdot}(\cdot)$ is non-smooth or non-gradient-Lipschitz, it is possible to move it into the corresponding independent term $f_j(\cdot)$ to ensure the required property of the coupling function $g(\cdot)$. This is for instance the case for the sparse penalization used over $\mathbf{H}$ which has been moved into $f_0(\cdot)$. Nonetheless, as mentioned above, the proximal operator associated with each $f_j(\cdot)$ is needed. Thus, even when the function consists of several terms, a closed-form expression of this operator should be known. Alternatively, one should be able to compose the proximal operators associated with each term of $f_j(\cdot)$ \cite{Yu2013}.


\begin{algorithm}[!ht]
\caption{PALM\label{alg:palm}}
\footnotesize
\SetAlgoLined
Initialize variables $\mathbf{H}^0$, $\boldsymbol{\theta}^0$, $\mathbf{Z}^0$, ${\mathbf{C}_{\mathcal{U}}}^0$ and $\mathbf{Q}^0$\;
Set $\alpha > 1$\;
\While{\text{stopping criterion not reached}}{
  $\mathbf{H}^{k+1} \in \prox{f_0}{\alpha L_\mathbf{H}}(\mathbf{H}^{k} - \frac{1}{\alpha L_\mathbf{H}} \nabla_\mathbf{H}g(\mathbf{H}^k,\boldsymbol{\theta}^k,\mathbf{Z}^k,{\mathbf{C}_{\mathcal{U}}^k},\mathbf{Q}^k))$\;
  $\boldsymbol{\theta}^{k+1} \in \prox{f_1}{\alpha L_{\boldsymbol{\theta}}}(\boldsymbol{\theta}^{k} - \frac{1}{\alpha L_{\boldsymbol{\theta}}} \nabla_{\boldsymbol{\theta}} g(\mathbf{H}^{k+1},\boldsymbol{\theta}^k,\mathbf{Z}^k,{\mathbf{C}_{\mathcal{U}}^k},\mathbf{Q}^k))$\;
  $\mathbf{Z}^{k+1} \in \prox{f_2}{\alpha L_\mathbf{Z}}(\mathbf{Z}^{k} - \frac{1}{\alpha L_\mathbf{Z}} \nabla_\mathbf{Z}g(\mathbf{H}^{k+1},\boldsymbol{\theta}^{k+1},\mathbf{Z}^k,{\mathbf{C}_{\mathcal{U}}^k},\mathbf{Q}^k))$\;
  $\mathbf{Q}^{k+1} \in \prox{f_3}{\alpha L_\mathbf{Q}}(\mathbf{Q}^{k} - \frac{1}{\alpha L_\mathbf{Q}} \nabla_\mathbf{Q}g(\mathbf{H}^{k+1},\boldsymbol{\theta}^{k+1},\mathbf{Z}^{k+1},{\mathbf{C}_{\mathcal{U}}}^k,\mathbf{Q}^k))$\;
  $\mathbf{C}_{\mathcal{U}}^{k+1} \in \prox{f_4}{\alpha L_\mathbf{\mathbf{C}_{\mathcal{U}}}}(\mathbf{C}_{\mathcal{U}}^{k} - \frac{1}{\alpha L_\mathbf{\mathbf{C}_{\mathcal{U}}}} \nabla_{\mathbf{C}_{\mathcal{U}}} g(\mathbf{H}^{k+1},\boldsymbol{\theta}^{k+1},\mathbf{Z}^{k+1},{\mathbf{C}_{\mathcal{U}}^k},\mathbf{Q}^{k+1}))$\;
}
\Return{$\mathbf{H}^{\textrm{end}},\boldsymbol{\theta}^{\textrm{end}},\mathbf{Z}^{\textrm{end}},\mathbf{Q}^{\textrm{end}},{\mathbf{C}_{\mathcal{U}}^{\textrm{end}}}$}
\end{algorithm}

\section{Application: hyperspectral images analysis}
\label{sec:hs-app}

A general framework has been introduced in the previous section. As an illustration, a particular instance of this generic framework is now considered, where explicit representation learning, classification and clustering are introduced. The specific case of hyperspectral images analysis is considered for this use case example.

Contrary to conventional color imaging which only captures the reflectance measure for three wavelengths (red, blue, green), hyperspectral imaging makes it possible to measure reflectance of the observed scene for several hundreds of wavelengths from visible to invisible domain. Each pixel of the image can thus be represented as a vector of reflectance, called spectrum, which characterizes the observed material.

One drawback of hyperspectral images is usually a weaker spatial resolution due to sensor limitations. The direct consequence of this poor spatial resolution is the presence of mixed pixels, i.e., pixels corresponding to areas containing several materials. Observed spectra are in this case the result of a specific mixture of the elementary spectra, called endmembers, associated with individual materials present in the pixel. The problem of retrieving the proportions of each material in each pixel is referred to as spectral unmixing~\cite{Bioucas-Dias2012}. This problem can be seen as a specific case of representation learning where the dictionary is composed of the set of endmembers standing for the endmember spectra and the coding matrix is the so-called abundance matrix containing the proportion of each material in each pixel.

Spectral unmixing is introduced as a representation learning task in Section~\ref{sec:unmixing}. The specific classifier used for this application is then explained in Section~\ref{sec:hs-classif} and finally Section~\ref{sec:hs-clust} presents the clustering adopted to relate the abundance matrix and the classification feature matrix.

\subsection{Spectral unmixing}
\label{sec:unmixing}

\begin{figure}
  \centering
  \includegraphics[width=0.8\columnwidth]{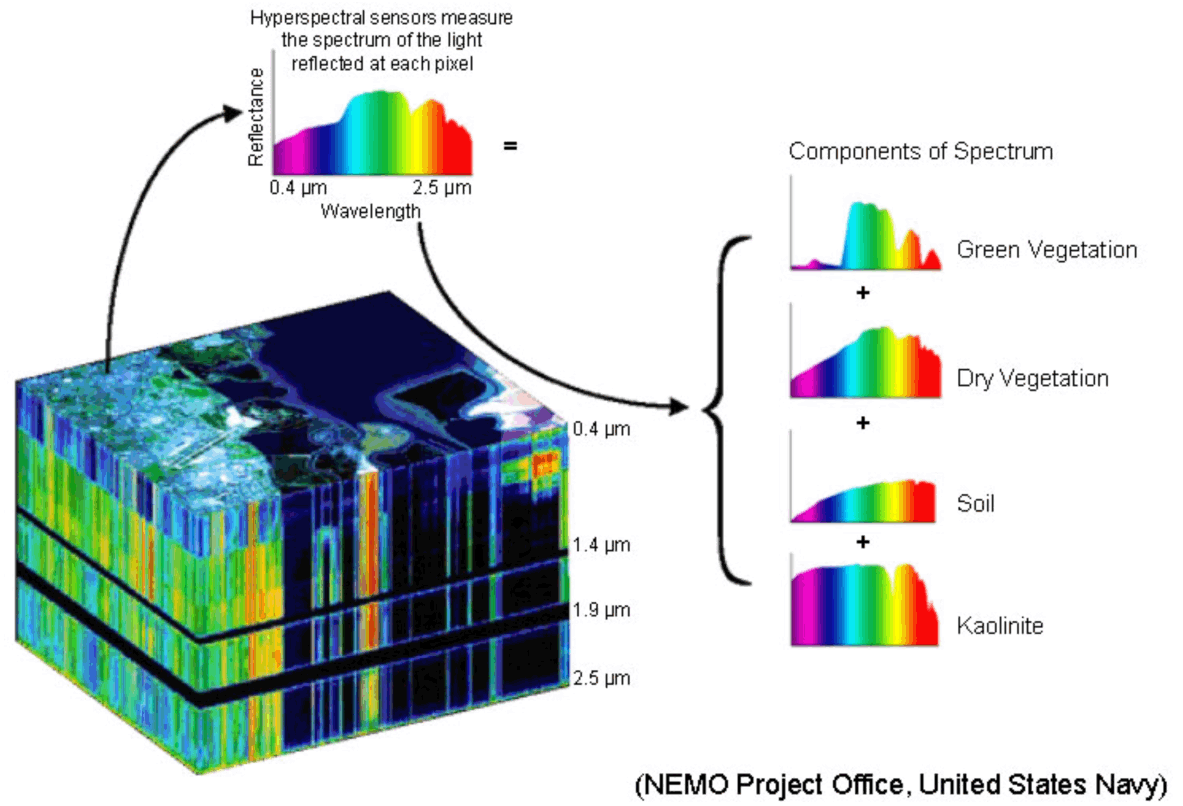}
  \caption{Spectral unmixing concept (source US Navy NEMO).\label{fig:unmixing}}
\end{figure}

As explained, each pixel of an hyperspectral image is characterized by a reflectance spectrum that physics theory approximates as a combination of endmembers, each corresponding to a specific material, as illustrated in Figure~\ref{fig:unmixing}. Formally, in this applicative scenario, the $L$-dimensional sample $\mathbf{y}_p$ denotes the $L$-dimensional spectrum of the $p$th pixel of the hyperspectral image ($p \in \mathcal{P}$). Each observation vectors $\mathbf{y}_p$ can be expressed as a function of the endmember matrix $\mathbf{W}$ (containing the $R$ elementary spectra) and the abundance vector $\mathbf{h}_p \in \mathbb{R}^R$ with $R \ll L$.

In the case of the most commonly adopted linear mixture model, each observation $\mathbf{y}_p$ is assumed to be a linear combination of the endmember spectra $\mathbf{w}_r$ ($r=1,\ldots,R$) corrupted by some noise, underlying the linear embedding \eqref{eq:psi}. Assuming a quadratic data-fitting term, the cost function associated with the representation learning task in \eqref{eq:repres-pb-gen} is written
\begin{equation}
\label{eq:unmix-cost}
  \costR(\mathbf{Y} | \mathbf{W} \mathbf{H}) = \frac{1}{2}\norm{\mathbf{Y} - \mathbf{W} \mathbf{H}}_{\mathrm{F}}^2.
\end{equation}

The abundance vector $\mathbf{h}_p$ is usually interpreted as a vector of proportions describing the proportion of each elementary component in the pixel. Thus, to derive an additive composition of the observed pixels, a nonnegative constraint is considered for each element of the abundance matrix $\mathbf{H}$, i.e., $\mathbb{H}=\mathbb{R}^{R\times P}_+$. In this work, no sum-to-one constraint is considered since it has been argued that leaving this constraint out offers a better adaptation to possible changes of illumination in the scene~\cite{Drumetz2016}. Additionally, as the endmember matrix $\mathbf{W}$ is the collection of reflectance spectra of the endmembers, it is also expected to be non-negative. When this dictionary needs to be estimated, the resulting problem is a sparse non-negative matrix factorization (NMF) task. When the dictionary is known or estimated beforehand, the resulting optimization problem is the nonnegative sparse coding problem
\begin{align}
\label{eq:unmix}
  &\min_{\mathbf{H} } \frac{1}{2} \norm{\mathbf{Y} - \mathbf{W}\mathbf{H}}_{\mathrm{F}}^2 + \lambda_{h} \norm{\mathbf{H}}_1 + \imath_{\mathbb{R}^{R\times P}_+}{(\mathbf{H})}
\end{align}
where the sparsity penalization actually supports the assumption that only a few materials are present in a given pixel.

\subsection{Classification}
\label{sec:hs-classif}

In the considered application, two loss functions associated with the classification problem have been investigated, namely quadratic loss and cross-entropy loss. One advantage of these two loss functions is that they can be used in a multi-class classification (i.e., with more than two classes). Moreover, this choice may fulfill the required conditions stated in Section~\ref{sec:optim} to apply PALM since, coupled with an appropriate $\phi(\cdot)$ function, both loss costs are smooth and gradient-Lipschitz according to each estimated variables.

\subsubsection{Quadratic loss}\label{sec:classif_quadratic}
The quadratic loss is the most simple way to perform a classification task and have been extensively used~\cite{Jiang2011,Zhang2001,Yang2011}. It is defined as
\begin{equation}
\label{eq:classif-cost-hyper}
  \costC(\mathbf{C} | \hat{\mathbf{C}}) = \frac{1}{2}\norm{\mathbf{C} \mathbf{D}- \hat{\mathbf{C}}\mathbf{D}}_{\mathrm{F}}^2
\end{equation}
where $\hat{\mathbf{C}}$ denotes the estimated attribution matrix. In \eqref{eq:classif-cost-hyper}, the $P\times P$ matrix $\mathbf{D}$ is introduced to weight the contribution of the labeled data with respect to the unlabeled one and to deal with the case of unbalanced classes in the training set. Weights are chosen to be inversely proportional to class frequencies in the input data. The weight matrix is defined as the diagonal matrix $\mathbf{D} = \mathrm{diag}[d_1,\ldots,d_P]$ with
\begin{equation}
  d_{p}=\left\{
    \begin{array}{ll}
      \sqrt{\frac{1}{|\mathcal{L}_i|}}, & \hbox{if $p\in\mathcal{L}_i$;} \\
      \sqrt{\frac{1}{|\mathcal{U}|}}, & \hbox{if $p\in\mathcal{U}$;}
    \end{array}
  \right.
\end{equation}
where $\mathcal{L}_i$ denotes the set of indexes of labeled pixels of the $i$th class ($i=1,\ldots,C$).
Thus, considering a linear classifier, the generic classification problem in \eqref{eq:classif-pb1} can be specified for the quadratic loss\vspace{-0.15cm}
\begin{equation}
  \min_{\mathbf{Q},\mathbf{C}_\mathcal{U}} \frac{1}{2}\norm{\mathbf{C}\mathbf{D} - \mathbf{Q}\mathbf{Z}\mathbf{D}}_{\mathrm{F}}^2  + \lambda_c \mathcal{R}_c(\mathbf{C}) + \imath_{\mathbb{S}_C^{|\mathcal{U}|}}(\mathbf{C}_\mathcal{U})
\end{equation}
where no additional constraints nor penalization is applied to the classifier parameters $\mathbf{Q}$. Besides, when samples obey a spatially coherent structure, as it is the case when analyzing hyperspectral images, it is often desirable to transfer this structure to the classification map. Such a characteristics can be achieved by considering a spatial regularization $\mathcal{R}_c(\mathbf{C})$ applied to the attributions vectors. Following this assumption, this work considers a regularized counterpart of the weighted vectorial total variation (vTV), promoting a spatially piecewise constant behavior of the classification map \cite{Liu2018}
\begin{equation}
\label{eq:regularization_TV}
  \norm{\mathbf{C}}_{\mathrm{vTV}} = \sum_{m,n} \beta_{m,n} \sqrt{\norm{\left[{\nabla}_{\mathrm{h}} \mathbf{C}\right]_{m,n}}_2^2 + \norm{\left[{\nabla}_{\mathrm{v}} \mathbf{C}\right]_{m,n}}_2^2 + \epsilon}
\end{equation}
where $(m,n)$ are the spatial position pixel indexes and $\left[{\nabla}_{\mathrm{h}}(\cdot)\right]_{m,n}$ and $\left[{\nabla}_{\mathrm{v}}(\cdot)\right]_{m,n}$ stand for horizontal and vertical discrete gradient operators evaluated at a given pixel\footnote{With a slight abuse of notations, $\mathbf{c}_{(m,n)}$ refers to the $p$th column of $\mathbf{C}$ where the $p$th pixel is spatially indexed by $(m,n)$.}, respectively, i.e.,\vspace{-0.15cm}
\begin{align*}
&\left[{\nabla}_{\mathrm{h}}\mathbf{C}\right]_{m,n} =\mathbf{c}_{(m+1,n)} - \mathbf{c}_{(m,n)}\\
&\left[{\nabla}_{\mathrm{v}}\mathbf{C}\right]_{m,n} =\mathbf{c}_{(m,n+1)} - \mathbf{c}_{(m,n)}.
\end{align*}
The weights $\beta_{m,n}$ can be computed beforehand to adjust the penalizations with respect to expected spatial variations of the scene. They can be estimated directly from the image to be analyzed or extracted from a complementary dataset as in~\cite{Uezato2018a}. They will be specified during the experiments reported in Section \ref{sec:exp}. Moreover, the smoothing parameter $\epsilon>0$ ensures the gradient-Lipschitz
property of the coupling term $g(\cdot)$, as required in Section \ref{sec:optim}.

\subsubsection{Cross-entropy loss} \label{sec:classif_entropy}
The quadratic loss has the advantage to be expressed simply and the associated Lipschitz constant of the partial gradients are trivially obtained. However, this loss function is known to be highly influenced by outliers which can result in a degraded predictive accuracy~\cite{Huber1964}. A more sophisticated way to conduct the classification task is to consider a cross-entropy loss
\begin{equation}
  \costC(\mathbf{C} | \hat{\mathbf{C}}) = -\sum_{p \in\mathcal{P}} d^2_p \sum_{i \in\mathcal{C}} c_{i,p} \log\left( \hat{c}_{i,p} \right)
\end{equation}
combined with a logistic regression, i.e., where the nonlinear mapping \eqref{eq:phi} is element-wise defined as
\begin{equation}
  \left[\phi\left(\mathbf{X}\right)\right]_{i,j} = \frac{1}{1+\exp(-x_{i,j})} =\sigm(x_{i,j})
\end{equation}
with $i\in\left\{1,\ldots,C\right\}$ and $p\in \mathcal{P}$. This classifier can actually be interpreted as a one-layer neural network with a sigmoid non-linearity. Cross-entropy loss is indeed a very conventional loss function in the neural network/deep learning community~\cite{Goodfellow2016}. In the present case, the corresponding optimization problem can be written\vspace{-0.15cm}
\begin{align}
\label{eq:classif-cost-hyper2}
  \min_{\mathbf{Q},\mathbf{C}_\mathcal{U}}& -\sum_{p \in\mathcal{P}} d_{p}^2 \sum_{i \in\mathcal{C}} c_{i,p} \log\left( \sigm(\mathbf{q}_{i:} \mathbf{z}_p)) \right) \nonumber\\
  &+ \lambda_q \mathcal{R}_q(\mathbf{Q}) + \lambda_c \norm{\mathbf{C}}_{\mathrm{vTV}} + \imath_{\mathbb{S}_C^{|\mathcal{U}|}}(\mathbf{C}_\mathcal{U})
\end{align}
where $\mathbf{q}_{i:} \in \mathbb{R}^{1\times K}$ denotes the $i$th line of the matrix $\mathbf{Q}$. The penalization $\mathcal{R}_q(\mathbf{Q})$ is here chosen as $\mathcal{R}_q(\mathbf{Q})=\frac{1}{2} \norm{\mathbf{Q}}_{\mathrm{F}}^2 $ to prevent the loss function to artificially decrease when $\norm{\mathbf{q}_{i:}}^2$ is increasing. This regularization has been extensively studied in the neural network literature where it is referred to as \emph{weight decay}~\cite{Goodfellow2016}. In \eqref{eq:classif-cost-hyper2}, the regularization $\mathcal{R}_c(\mathbf{C}_{\mathcal{U}})$ applied to the attribution matrix is chosen again as a vTV-like penalization (see \eqref{eq:regularization_TV}).

\subsection{Clustering}
\label{sec:hs-clust}

For the considered application, the conventional $k$-means algorithm has been chosen because of its straightforward formulation as an optimization problem.
By denoting $\boldsymbol{\theta} = \{\mathbf{B}\}$ a $R\times K$ matrix collecting $K$ centroids, the clustering task \eqref{eq:clust-pb} can be rewritten as the following NMF problem \cite{Pompili2014} \vspace{-0.15cm}
\begin{equation}
\label{eq:clust-cstr-pb}
  \min_{\mathbf{Z},\mathbf{B}} \frac{1}{2}\norm{\mathbf{H} - \mathbf{B} \mathbf{Z}}_{\mathrm{F}}^2 + \lambda_z \mathcal{R}_z(\mathbf{Z}) +\imath_{\mathbb{S}_K^P}(\mathbf{Z})  + \imath_{\mathbb{R}_+^{R\times K}}(\mathbf{B})
\end{equation}
where $\mathcal{R}_z(\mathbf{Z})$ should promote $\mathbf{Z}$ to be composed of orthogonal lines. Combined with the nonnegativity and sum-to-one constraints, it would ensure that $\mathbf{z}_p$ is a vector of zeros except for its $k$th component equal to $1$, i.e., meaning that the $p$th pixel belongs to the $k$th cluster. However, handling this orthogonality property within the PALM optimization scheme detailed in Section \ref{sec:optim} is not straightforward, in particular because the proximal operator associated to this penalization cannot be explicitly computed. In this work, we propose to remove this orthogonality constraint since relaxed attribution vectors may be richer feature vectors for the classification task.

\subsection{Multi-objective problem}
\label{sec:hs-global-pb}

Based on the quadratic and cross-entropy loss functions considered in the classification task, two distinct global optimization problems are obtained. When considering the quadratic loss of Section \ref{sec:classif_quadratic}, the multi-objective problem \eqref{eq:glob-pb} writes\vspace{-0.15cm}
\begin{align}
\label{eq:hs-glob-pb-quad}
  &\min_{\substack{ \mathbf{H},\mathbf{Q},\mathbf{Z}\\ \mathbf{C}_{\mathcal{U}},\mathbf{B} }}
    \frac{\lambda_0}{2} \norm{\mathbf{Y} - \mathbf{W}\mathbf{H}}_{\mathrm{F}}^2 + \lambda_{h} \norm{\mathbf{H}}_1 + \imath_{\mathbb{R}_+^{R\times P}}(\mathbf{H}) \nonumber\\
  &+ \frac{\lambda_1}{2} \norm{\mathbf{C}\mathbf{D} - \mathbf{Q}\mathbf{Z}\mathbf{D}}_{\mathrm{F}}^2 + \lambda_{c} \norm{\mathbf{C}}_{\mathrm{vTV}} +  \imath_{\mathbb{S}_C^{|\mathcal{U}|}}(\mathbf{C}_\mathcal{U}) \nonumber\\
  &+ \frac{\lambda_2}{2} \norm{\mathbf{H} - \mathbf{B}\mathbf{Z}}_F^2 +\imath_{\mathbb{S}_K^P}(\mathbf{Z})  + \imath_{\mathbb{R}_+^{R\times K}}(\mathbf{B}).
\end{align}
Instead, when considering the cross-entropy loss function proposed in Section \ref{sec:classif_entropy}, the optimization problem \eqref{eq:glob-pb} is defined as\vspace{-0.15cm}
\begin{align}
\label{eq:hs-glob-pb-ce}
&\min_{\substack{ \mathbf{H},\mathbf{Q},\mathbf{Z}\\ \mathbf{C}_{\mathcal{U}},\mathbf{B} }}
    \frac{\lambda_0}{2} \norm{\mathbf{Y} - \mathbf{W}\mathbf{H}}_{\mathrm{F}}^2 + \lambda_{h} \norm{\mathbf{H}}_1 + \imath_{\mathbb{R}_+^{R\times P}}(\mathbf{H}) \nonumber\\
  & - \frac{\lambda_1}{2} \sum_{p \in\mathcal{P}} d_{p}^2 \sum_{i \in\mathcal{C}} c_{i,p} \log\left( \sigm(-\mathbf{q}_{i:} \mathbf{z}_p)) \right) \nonumber\\
  &+ \frac{\lambda_q}{2} \norm{\mathbf{Q}}_{\mathrm{F}}^2 + \lambda_{c} \norm{\mathbf{C}}_{\mathrm{vTV}} +  \imath_{\mathbb{S}_C^{|\mathcal{U}|}}(\mathbf{C}_\mathcal{U}) \nonumber\\
  &+ \frac{\lambda_2}{2} \norm{\mathbf{H} - \mathbf{B}\mathbf{Z}}_F^2 +\imath_{\mathbb{S}_K^P}(\mathbf{Z})  + \imath_{\mathbb{R}_+^{R\times K}}(\mathbf{B}).
%
\end{align}
Both problems are particular instances of nonnegative matrix co-factorization~\cite{Yokoya2012,Yoo2010}. To summarize, the hyperspectral pixel is first described as a combination of elementary spectra through the learning representation step, aka spectral unmixing. Then, assuming that there exist groups of pixels resulting from the same mixture of materials, a clustering is performed among the abundance vectors. And finally, attribution vectors to the clusters are used as feature vectors for the classification supporting the idea that classes are made of a mixture of clusters. For both multi-objective problems \eqref{eq:hs-glob-pb-quad} and \eqref{eq:hs-glob-pb-ce}, all conditions required to the use of PALM algorithm described in Section~\ref{sec:optim} are met. Details regarding the two optimization schemes dedicated to these two problems are reported in the Appendix.

\subsection{Complexity analysis}

Regarding the computational complexity of the proposed Algorithm~\ref{alg:palm}, deriving the gradients shows that it is dominated by matrix product operations. It yields that the algorithm has an overall computational cost in $\mathcal{O}(NK^2P)$ where $N$ is the number of iterations.

\section{Experiments}
\label{sec:exp}

\subsection{Implementation details}
\label{sec:hs-impl}

Before presenting the experimental results, it is worth clarifying the choices which have been made regarding the practical implementation of the proposed algorithms for the considered application. Important aspects are discussed below.\\

\begin{figure}[!ht]
  \centering
    \begin{tikzpicture}
      \begin{semilogyaxis}[width=0.9\columnwidth,legend style={font=\scriptsize,at={(1.,1.05)},anchor=north east},grid=both,grid style={line width=.1pt, draw=gray!30},major grid style={line width=.2pt,draw=gray!60},axis x line=left,axis y line=left,xlabel={Iteration},ylabel={Value},ylabel style={align=center},font=\footnotesize]
        \addplot+[thick] table[x expr=\coordindex,y index=0] {Fig/objFct_convergence.txt};
        \addplot+[thick] table[x expr=\coordindex,y index=1] {Fig/objFct_convergence.txt};
        \addplot+[thick] table[x expr=\coordindex,y index=2] {Fig/objFct_convergence.txt};
        \addplot+[thick] table[x expr=\coordindex,y index=3] {Fig/objFct_convergence.txt};
        \addplot+[very thick,black] table[x expr=\coordindex,y index=4] {Fig/objFct_convergence.txt};
        \legend{Repres.,Cluster.,Classif.,vTV,Overall}
      \end{semilogyaxis};
    \end{tikzpicture}
  \caption{Convergence of the various terms of objective function (representation learning, clustering, classification, vTV, total).\label{fig:convergence}}
\end{figure}
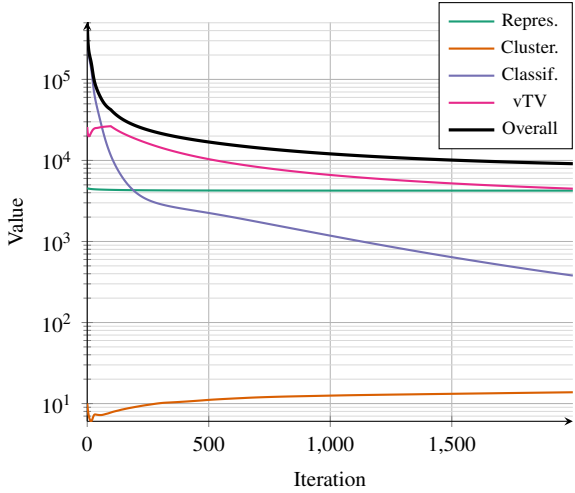

\noindent\textbf{Convergence diagnosis and stopping rule --}  In all experiments conducted hereafter, the value of the objective function is monitored at each iteration to determine if convergence has been reached. The normalized difference between the last two consecutive values of the objective function is compared to a threshold and the algorithm stops when the criterion is smaller than this threshold (set as $10^{-4}$ for the conducted experiments). Figure~\ref{fig:convergence} shows one example of the behavior of the objective function along the iterations as well as the behavior of several terms composing this overall objective function. As it can be observed from the figure, the global objective function is decreasing over the iteration, which is theoretically ensured by the PALM algorithm.\\

\noindent\textbf{Initialization --} As PALM algorithm only ensures convergence to a critical point and not a global optimum, it remains sensitive to initialization, which needs to be carefully chosen to reach relevant solutions. The initialization of the parameters associated with the learning representation and clustering steps relies on the self-dictionary learning method proposed in \cite{Gillis2018}. This method proposes to use observed pixels of the image as dictionary elements. The underlying assumption is that the image contains pure pixels, i.e., composed of only a single material. Formally, the initial estimate $\mathbf{H}^0$ of $\mathbf{H}$ is chosen as\vspace{-0.15cm}
\begin{equation}
\label{eq:pb-self-dict}
  \mathbf{H}^0 = \argmin_{\mathbf{H}} \frac{1}{2} \norm{\mathbf{Y} - \tilde{\mathbf{Y}}\mathbf{H}}_{\mathrm{F}}^2 + \alpha \norm{\mathbf{H}}_{1,2}\vspace{-0.15cm}
\end{equation}
where $\norm{\mathbf{H}}_{1,2} = \sum_{r=1}^{R} \norm{\mathbf{h}_{r,:}}_2$ promotes the use of a reduced number of pixels as dictionary elements and $\tilde{\mathbf{Y}}$ is a submatrix of $\mathbf{Y}$ containing the pixel candidates to be used as dictionary elements. Following the strategy similarly proposed in \cite{Gillis2018}, this subset $\tilde{\mathbf{Y}}$ is built as follows: $i$) for each class of the training set, a $k$-means is applied to the labeled samples to identify $J$ clusters, $ii$) within a given class, one candidate is retained from each cluster as the pixel the farthest away from the centers of the other clusters (in term of spectral angle distance). This procedure provides a subset $\tilde{\mathbf{Y}}$ composed of $J\times C$ spectrally diverse candidates extracted from the labeled samples.

Then, regarding the representation learning step, only active elements in $\tilde{\mathbf{Y}}$, i.e., those associated with non-zero rows in $\mathbf{H}^0$, are kept to define the dictionary $\mathbf{W}$. Finally, to initialize the variables involved in the clustering step, a $k$-means is conducted on $\mathbf{H}^0$ and the identified centroids are chosen as $\mathbf{B}^0$ while the corresponding attribution vectors define $\mathbf{Z}^0$. Finally, the classification parameters $\mathbf{Q}^0$ and attribution vectors $\mathbf{C}_{\mathcal{U}}^0$ are randomly initialized.\\

\noindent\textbf{Weighting the vTV --} As explained in Section~\ref{sec:classif-pb}, the classification is regularized by a weighted smooth vTV regularization. When all not fixed to the same value, the weights offer the possibility to account for natural boundaries in the observed scene, i.e., variations in the classification map are expected to be localized at the edges in the image. As in~\cite{Uezato2018a}, an auxiliary dataset informing about the spatial structure of the image can be used to adjust these weights. Instead, in this work, we assume that no such external information is available. Thus these weights are directly computed from the hyperspectral image. More precisely, a virtually observed panchromatic image $\mathbf{y}_{\mathrm{PAN}} \in \mathbb{R}^P$, i.e. a single band image, is first synthetized by averaging the bands of the hyperspectral image $\mathbf{Y}$. Then, the weights are chosen as\vspace{-0.15cm}
\begin{equation}
\beta_{m,n} = \frac{\tilde{\beta}_{m,n}}{\sum_{p,q}  \tilde{\beta}_{p,q}}
\
\text{with}
\
\tilde{\beta}_{m,n} = \frac{1}{ \norm{\left[\nabla \mathbf{y}_{\mathrm{PAN}}\right]_{m,n}}_2 + \sigma }\vspace{-0.15cm}
\end{equation}
where $\nabla(\cdot)  = \left[\nabla_{\mathrm{h}}(\cdot)\  \nabla_{\mathrm{v}}(\cdot)\right]^T $ is the gradient operator and $\sigma$ is an hyperparameter chosen as $\sigma=0.01$ to avoid numerical problems and to control the adaptive weighting (the larger $\sigma$, the less variation in the weighting)~\cite{Strong1997}.\\

\noindent\textbf{Hyperparameter scaling --} To balance the size and the dynamics of the matrices involved in the cofactorization problem, the hyperparameters $\lambda_0$ and $\lambda_q$ in \eqref{eq:hs-glob-pb-quad} and \eqref{eq:hs-glob-pb-ce} have been set as
\begin{equation}
\label{eq:hyparameters}
\lambda_0 = \frac{1}{L\norm{\mathbf{Y}}_{\infty}^2} {\tilde{\lambda}_0}, \quad
\lambda_q =  \frac{P}{C}{\tilde{\lambda}_q}.
\end{equation}
Then, for each experiment presented hereafter, the parameters  $\tilde{\lambda}_{\cdot}$ have been empirically adjusted to obtain consistent results.

\begin{figure}[!ht]
  \centering
  \begin{tabular}{@{}cc}
    \includegraphics[width=0.45\columnwidth]{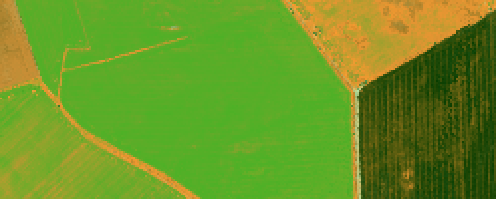}&
    \includegraphics[width=0.45\columnwidth]{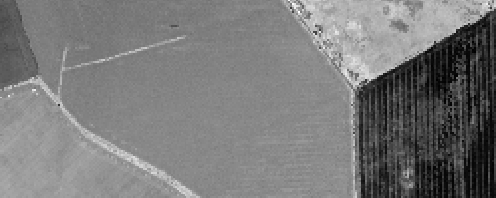}\\
    (a) & (b) \\
    \includegraphics[width=0.45\columnwidth]{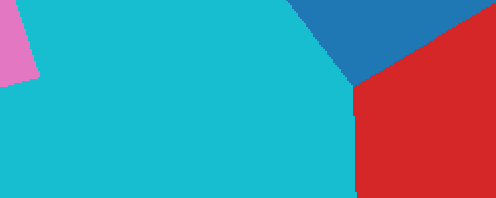}&
    \includegraphics[width=0.45\columnwidth]{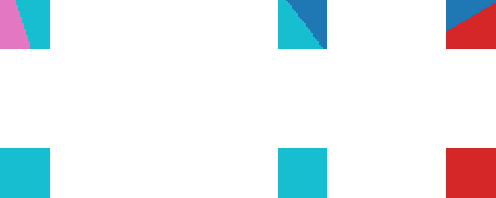}\\
    (c) & (d)
  \end{tabular}
  \caption{Synthetic image: (a) colored composition of the hyperspectral image $\mathbf{Y}$, (b) panchromatic image $\mathbf{y}_{\mathrm{PAN}}$, (c) classification ground-truth, (d) training set.\label{fig:semisynth-data}}
\end{figure}

\subsection{Synthetic hyperspectral image}
\label{sec:synth-exp}

\noindent{\textbf{Data generation --}} First, to assess the relevance of the proposed model, experiments have been conducted on synthetic images. These synthetic images have been generated using a real hyperspectral image which has been unmixed using the well-established unmixing method SUnSAL~\cite{Bioucas-Dias2010}. The extracted abundance maps and a set of $6$ pure spectra from the hyperspectral library ASTER have been used to build a synthetic hyperspectral images with a realistic spatial organization. The resulting $100$-by-$250$ pixel image presented in Figure~\ref{fig:semisynth-data} is composed of $L=385$ spectral bands. The image is associated with a classification groundtruth ($C=4$) based on the groundtruth of the original real image and a subpart of this groundtruth is assumed known and therefore used as training dataset for the supervised classification step.

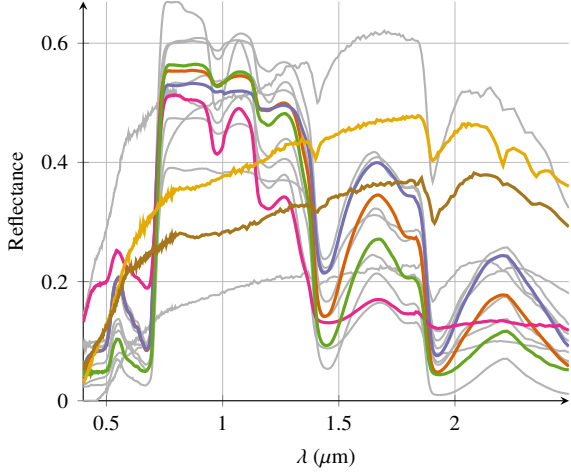
\begin{figure}[!ht]
  \centering
    \begin{tikzpicture}
      \begin{axis}[width=0.9\columnwidth,ymin=0.,grid,axis x line=left,axis y line=left,xlabel={$\lambda$ ($\mu$m)},ylabel={Reflectance},ylabel style={align=center},font=\footnotesize]
        \addplot[thick,gray!60] table[x index=0, y expr=0.01*\thisrowno{3}] {Fig/semisynth_spectra.txt};
        \addplot[thick,gray!60] table[x index=0, y expr=0.01*\thisrowno{4}] {Fig/semisynth_spectra.txt};
        \addplot[thick,gray!60] table[x index=0, y expr=0.01*\thisrowno{5}] {Fig/semisynth_spectra.txt};
        \addplot[thick,gray!60] table[x index=0, y expr=0.01*\thisrowno{7}] {Fig/semisynth_spectra.txt};
        \addplot[thick,gray!60] table[x index=0, y expr=0.01*\thisrowno{9}] {Fig/semisynth_spectra.txt};
        \addplot[thick,gray!60] table[x index=0, y expr=0.01*\thisrowno{10}] {Fig/semisynth_spectra.txt};
        \addplot[thick,gray!60] table[x index=0, y expr=0.01*\thisrowno{11}] {Fig/semisynth_spectra.txt};
        \addplot[thick,gray!60] table[x index=0, y expr=0.01*\thisrowno{14}] {Fig/semisynth_spectra.txt};
        \addplot[thick,gray!60] table[x index=0, y expr=0.01*\thisrowno{15}] {Fig/semisynth_spectra.txt};
        \addplot+[very thick] table[x index=0, y expr=0.01*\thisrowno{1}] {Fig/semisynth_spectra.txt};
        \addplot+[very thick] table[x index=0, y expr=0.01*\thisrowno{2}] {Fig/semisynth_spectra.txt};
        \addplot+[very thick] table[x index=0, y expr=0.01*\thisrowno{6}] {Fig/semisynth_spectra.txt};
        \addplot+[very thick] table[x index=0, y expr=0.01*\thisrowno{8}] {Fig/semisynth_spectra.txt};
        \addplot+[very thick] table[x index=0, y expr=0.01*\thisrowno{12}] {Fig/semisynth_spectra.txt};
        \addplot+[very thick] table[x index=0, y expr=0.01*\thisrowno{13}] {Fig/semisynth_spectra.txt};
      \end{axis};
    \end{tikzpicture}
  \caption{Spectra used as dictionary $\mathbf{W}$. The $6$ color spectra have been used to generate the semi-synthetic image (4 vegetation spectra and 2 soil spectra).\label{fig:semisynth-spectra}}
\end{figure}

Moreover, in this experiment, the endmember matrix $\mathbf{W}$ comprises the $6$ spectra actually used to generate the image. To evaluate the robustness of the method in a challenging scenario, these $6$ initial endmember spectra are complemented with $9$ endmembers not present in the image but very correlated with the $6$ actually used ones. The endmember matrix is thus composed of $R=15$ spectra depicted in Figure~\ref{fig:semisynth-spectra}.\\

\begin{table*}[!ht]
  \centering
  \caption{Synthetic data: unmixing and classification results.\label{tab:res-synth}}
  \begin{tabular}{lccccc}\toprule
    Model & F1-mean & Kappa & RMSE$(\hat{\mathbf{H}})$ & RE & Time (s) \\
    \midrule
    Cofact-Q    & $0.911$ ($\pm 3.5\times 10^{-3}$) & $0.893$ ($\pm 3.5\times 10^{-3}$) & $0.0528$ ($\pm 1.1\times 10^{-4}$)   & $0.32$ ($\pm 8.9\times 10^{-4}$)  & $80$ ($\pm 6$) \\
    Cofact-CE   & $0.899$ ($\pm 5.4\times 10^{-2}$) & $0.880$ ($\pm 6.2\times 10^{-2}$) & $0.0524$ ($\pm 1.3\times 10^{-4}$)   & $0.27$ ($\pm 2.2\times 10^{-3}$)  & $61$ ($\pm 4$) \\
    MLR         & $0.873$ ($\pm 2.6\times 10^{-3}$) & $0.882$ ($\pm 2.3\times 10^{-3}$) & N$\backslash$A            & N$\backslash$A         & $92$ ($\pm 14$) \\
    RF          & $0.913$ ($\pm 1.4\times 10^{-3}$) & $0.907$ ($\pm 1.3\times 10^{-4}$) & N$\backslash$A            & N$\backslash$A         & $0.9$ ($\pm 0.08$) \\
    ResNet      & $0.913$ ($\pm 1.6\times 10^{-2}$) & $0.943$ ($\pm 4.6\times 10^{-3}$) & N$\backslash$A            & N$\backslash$A         & $220$ ($\pm 12$)$^{*}$ \\
    SSFPCA+SVM  & $0.918$ ($\pm 8.3\times 10^{-4}$) & $0.911$ ($\pm 2.4\times 10^{-3}$) & N$\backslash$A            & N$\backslash$A         & $4.0$ ($\pm 0.05$) \\
    \textsc{fc}-SUnSAL+MLR          & $0.893$ ($\pm 6.4\times 10^{-4}$) & $0.912$ ($\pm 3.7\times 10^{-4}$) & $0.120$  ($\pm 3.1\times 10^{-6}$)   & $0.37$ ($\pm 5.1\times 10^{-5}$)  & $6$ ($\pm 0.3$) \\
    \textsc{csr}-SUnSAL+MLR         & $0.888$ ($\pm 1.0\times 10^{-3}$) & $0.911$ ($\pm 5.0\times 10^{-4}$) & $0.125$  ($\pm 3.0\times 10^{-6}$)   & $0.36$ ($\pm 4.2\times 10^{-5}$)  & $9$ ($\pm 0.5$) \\
    D-KSVD        & $0.520$ ($\pm 3.1\times 10^{-3}$) & $0.653$ ($\pm 3.4\times 10^{-2}$) & N$\backslash$A            & $0.23$ ($\pm 4.1\times 10^{-2}$)  & $382$ ($\pm 9$) \\
    LC-KSVD     & $0.879$ ($\pm 3.7\times 10^{-4}$) & $0.904$ ($\pm 1.0\times 10^{-4}$) & N$\backslash$A            & $30.4$ ($\pm 1.0\times 10^{-4}$)  & $96$ ($\pm 1$) \\
    \bottomrule
    \multicolumn{6}{l}{\footnotesize $^{*}$Based on a GPU implementation run on a computer cluster.} \\
  \end{tabular}
\end{table*}

\begin{figure*}[!ht]
  \centering
  \begin{tabular}{@{}cccccc}
    \includegraphics[width=0.3\columnwidth]{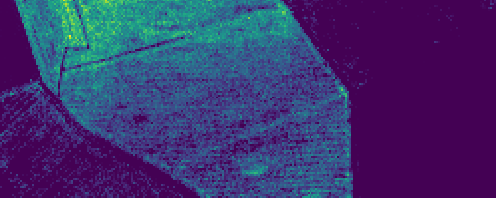}&
    \includegraphics[width=0.3\columnwidth]{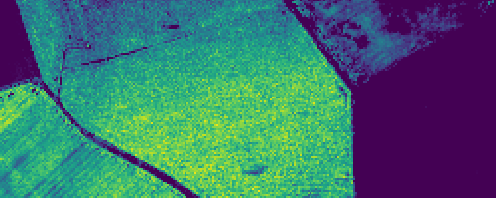}&
    \includegraphics[width=0.3\columnwidth]{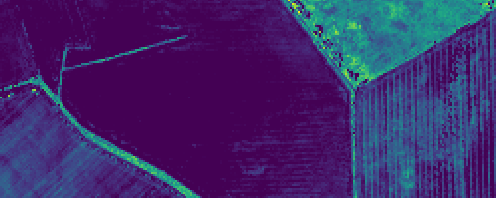}&
    \includegraphics[width=0.3\columnwidth]{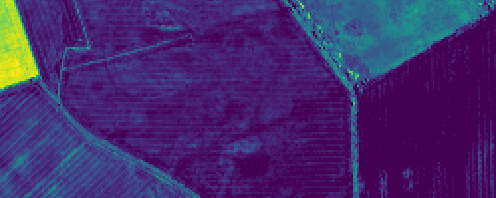}&
    \includegraphics[width=0.3\columnwidth]{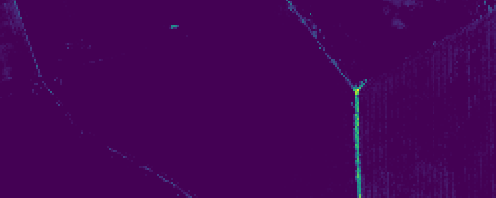}&
    \includegraphics[width=0.3\columnwidth]{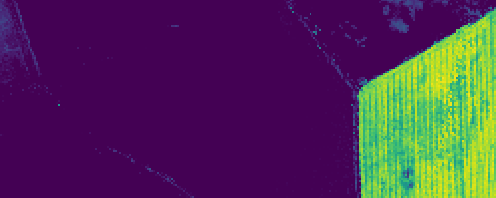}\\
    \includegraphics[width=0.3\columnwidth]{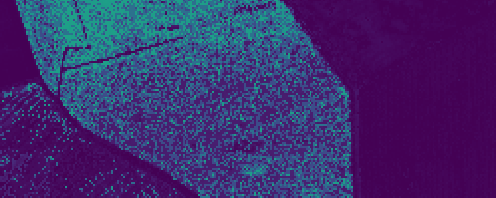}&
    \includegraphics[width=0.3\columnwidth]{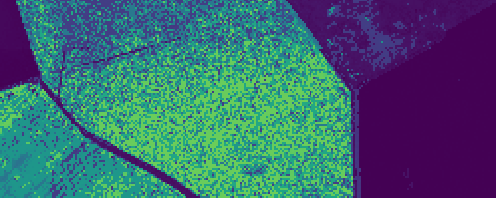}&
    \includegraphics[width=0.3\columnwidth]{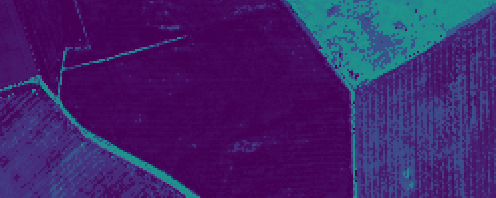}&
    \includegraphics[width=0.3\columnwidth]{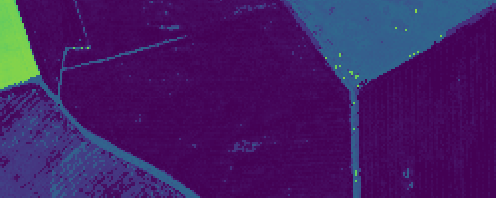}&
    \includegraphics[width=0.3\columnwidth]{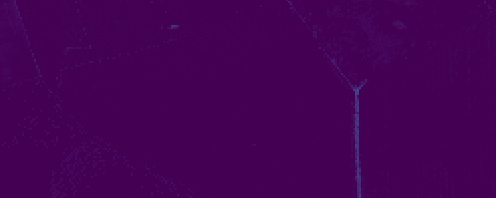}&
    \includegraphics[width=0.3\columnwidth]{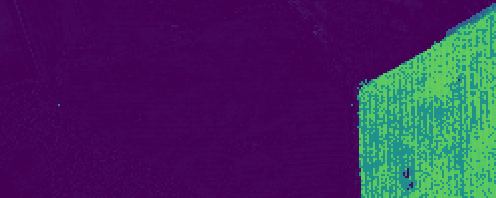}\\
    \includegraphics[width=0.3\columnwidth]{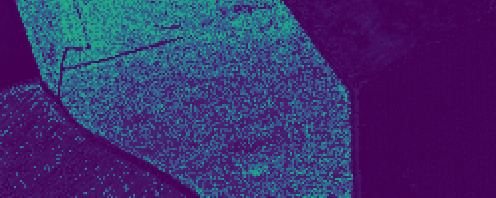}&
    \includegraphics[width=0.3\columnwidth]{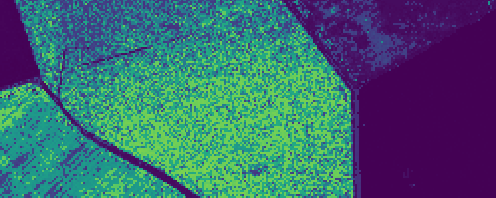}&
    \includegraphics[width=0.3\columnwidth]{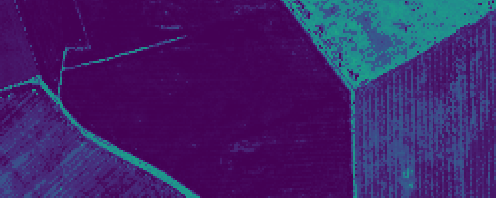}&
    \includegraphics[width=0.3\columnwidth]{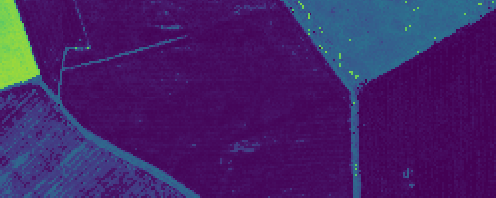}&
    \includegraphics[width=0.3\columnwidth]{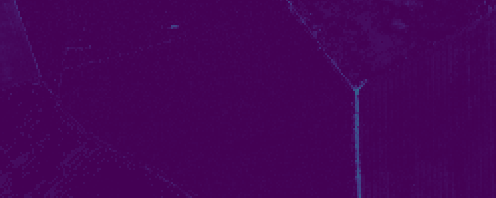}&
    \includegraphics[width=0.3\columnwidth]{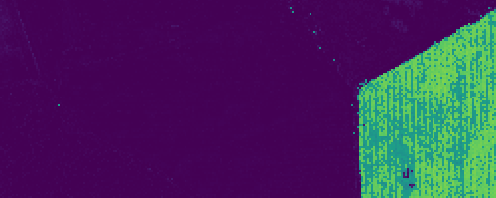}\\
    \includegraphics[width=0.3\columnwidth]{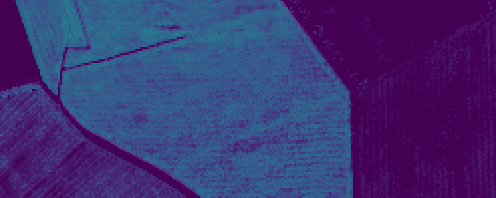}&
    \includegraphics[width=0.3\columnwidth]{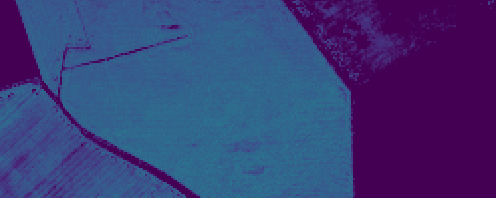}&
    \includegraphics[width=0.3\columnwidth]{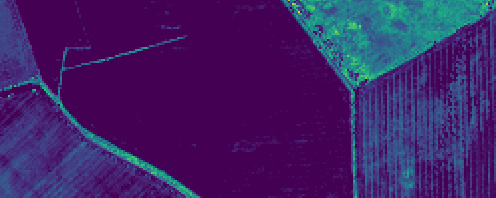}&
    \includegraphics[width=0.3\columnwidth]{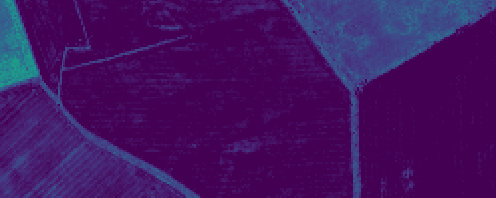}&
    \includegraphics[width=0.3\columnwidth]{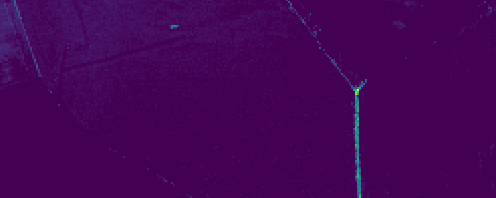}&
    \includegraphics[width=0.3\columnwidth]{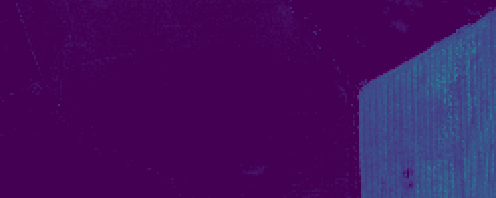}\\
    \includegraphics[width=0.3\columnwidth]{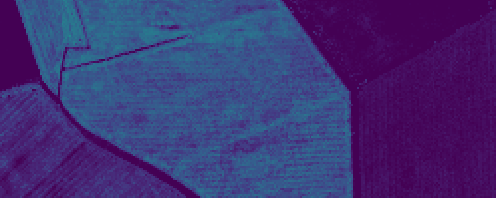}&
    \includegraphics[width=0.3\columnwidth]{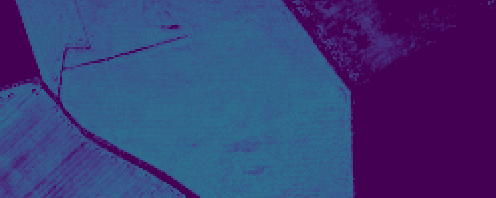}&
    \includegraphics[width=0.3\columnwidth]{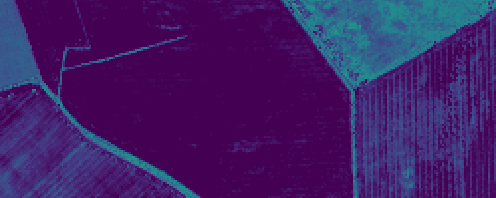}&
    \includegraphics[width=0.3\columnwidth]{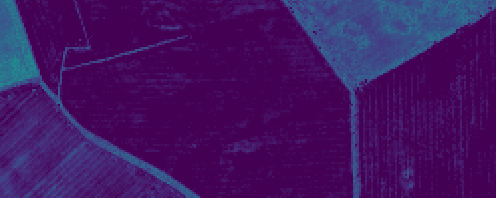}&
    \includegraphics[width=0.3\columnwidth]{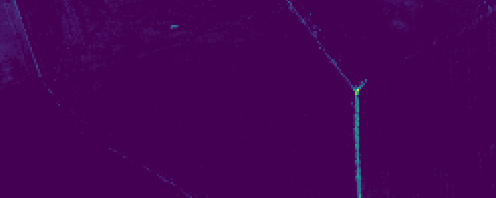}&
    \includegraphics[width=0.3\columnwidth]{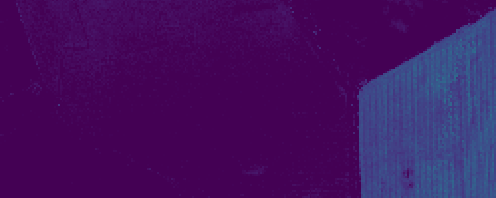}\\
  \end{tabular}
  \caption{Synthetic data: abundance maps of the $6$ actual endmembers (from left to right): ($1$st row) ground-truth, ($2$nd row) Cofact-Q, ($3$rd row) Cofact-CE, ($4$rd row) \textsc{fc}-SUnSAL and ($5$th row) \textsc{csr}-SUnSAL.\label{fig:semisynth-abund}}
\end{figure*}

\noindent{\textbf{Compared methods --}} The proposed methods with quadratic (Q) and cross-entropy (CE) classification losses, denoted respectively by Cofact-Q and Cofact-CE, have been compared with state-of-the-art classification and unmixing methods. First, one considered competing method is the random forest (RF) classifier, which has been extensively used for the hyperspectral image classification. \revBis{Then, the convolutional neural network (CNN) proposed in~\cite{Paoletti2019} has also been tested. This CNN architecture, referred to as ResNet, is based on a residual network specifically designed for hyperspectral image classification.} \rev{Additionally, the performance of the classification method proposed in~\cite{uddin2019} has been evaluated. This method, referred to as SSFPCA+SVM, relies on a so-called spectrally-segmented folded PCA (SSFPCA) as a feature extraction step, followed by a RBF-kernel SVM classifier.} \revBis{Finally, a multinomial logistic regression classifier (MLR) has also been applied directly on the observations. This classifier is equivalent to the classification term proposed in the Cofact-CE method. Thus it will illustrate the interest of using a representation learning  step before performing the classification.} \rev{Parameters of the RF and the SVM have been adjusted using cross-validation with a grid-search strategy and we used the implementations provided in the \emph{scikit-learn} Python library \cite{scikit-learn}.} \rev{The parameters of SSFPCA have been set based on the study provided in the original paper.} \revBis{The implementation and parameters proposed by the authors has been used for the ResNet method. All methods except ResNet have been run on a desktop computer with $16$Gb of RAM and Intel(R) Xeon(R) CPU E5-1630 v4 @ $3.70$GHz$\times8$ processor. Due to its high computational load, the ResNet method has been run on a DELL T630 server with $2$ Intel(R) Xeon(R) CPU 2640 v4, $2\times100$Gb of RAM and a Nvidia GTX 1080 TI GPU.}  

Besides, two unmixing methods proposed in~\cite{Bioucas-Dias2010} has been tested, namely the fully constrained least squares (\textsc{fc}-SUnSAL) and the constrained sparse regression (\textsc{csr}-SUnSAL). \textsc{fc}-SUnSAL basically relies on the same data fitting term \eqref{eq:unmix-cost} considered in the proposed cofactorization method, under non-negativity and sum-to-one constraints applied to the abundance vectors. Conversely, the \textsc{csr}-SUnSAL problem removes the sum-to-one constraint and introduces a $\ell_1$-norm penalization on the abundance vectors. It thus solves \eqref{eq:unmix} where the associated regularization parameter $\lambda_h$ is tuned using a grid-search strategy. These two methods use an augmented Lagrangian splitting algorithm to recover the abundance vectors. Additionally, these abundance vectors are subsequently used as input features of a MLR classifier. \rev{This classifier is linear and its combination with the \textsc{csr}-SUnSAL unmixing algorithm, referred to as \textsc{csr}-SUnSAL+MLR, yields a sequential counterpart of the proposed Cofact-CE method. In particular, comparing the resulting classification performance with the performance of Cofact-CE allows the benefit of introducing the clustering coupling term to be assessed.}

Besides, the proposed method has been also compared with the discriminative K-SVD (D-KSVD) method proposed in~\cite{Zhang2010}. The D-KSVD problem has strong similarities with the proposed cofactorization problem. Indeed, it corresponds to a $\ell_0$-penalized representation learning and a classification with a quadratic loss. It aims at learning a dictionary suitable for the classification problem and performs a linear classification on the coding vectors. For this reason, the dictionary $\mathbf{W}$ is only used as an initialization for D-KSVD, while it remains fixed for the unmixing and proposed cofactorization methods. Similarly, the label consistent K-SVD (LC-KSVD) is also considered \cite{Jiang2011}. This model has been proposed as an improvement of D-KSVD where an additional term ensures that the dictionary elements are class-specific. Hyperparameters of D-KSVD and LC-KSVD have been manually adjusted in order to get the best results.
When implementing the PALM algorithm proposed in Section \ref{sec:optim}, the normalized regularization parameters in \eqref{eq:hyparameters} have been fixed as $\tilde{\lambda}_0 = 100$, $\lambda_1 = \lambda_2 = 1$, $\lambda_h = \lambda_q = 0.1$ and $\tilde{\lambda}_{c} = 10^{-3}$. Finally, the number of clusters has been set to $K=10$. \revTer{The influence of these parameters are empirically studied in the associated companion report \cite{Lagrange2019TR}.}\\

\noindent{\textbf{Figure-of-merits --}} Several metrics are computed to quantify the quality of the classification and unmixing tasks. For classification, two widely-used metrics are used, namely Cohen's kappa and the averaged F1-score over all classes \cite{Congalton2008}. For unmixing, reconstruction error (RE) and root global mean squared error (RMSE) are computed as follows\vspace{-0.10cm}
\begin{align}
\label{eq:metric}
  &\mathrm{RE} = \sqrt{\frac{1}{PL} \norm{\mathbf{Y}-\mathbf{W}\hat{\mathbf{H}}}_{\mathrm{F}}^2}, \nonumber \\
  &\mathrm{RMSE}(\hat{\mathbf{H}}) = \sqrt{\frac{1}{PR} \norm{\mathbf{H}_{\mathrm{true}}-\hat{\mathbf{H}}}_{\mathrm{F}}^2}\vspace{-0.15cm}
\end{align}
where $\mathbf{H}_{\mathrm{true}}$ and $\hat{\mathbf{H}}$ are the actual and estimated abundance matrices. All these performance metrics are complemented with the computational times. Again, note that for all methods, similar computational framework have been considered except for the CNN-based algorithm whose complexity requires a specific GPU implementation embedded on a computer cluster.\\

\begin{figure}[!ht]
  \centering
  \begin{tabular}{@{}c@{~}c@{~}c}
    \includegraphics[width=0.32\columnwidth]{Fig/Res-synth/semisynthAisa_GT.png}&
    \includegraphics[width=0.32\columnwidth]{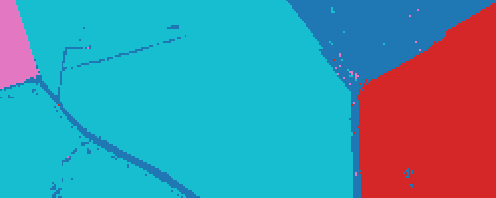}&
    \includegraphics[width=0.32\columnwidth]{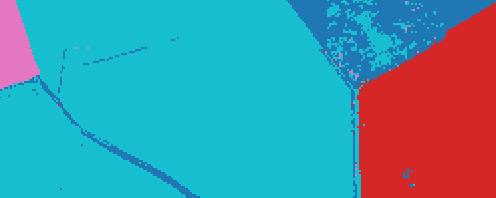}\\
    (a) & (b) & (c) \\
    \includegraphics[width=0.32\columnwidth]{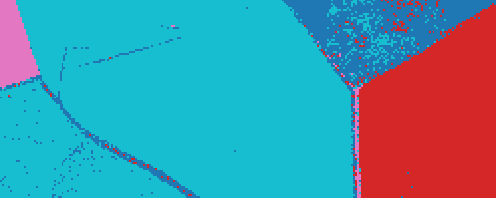}&
    \includegraphics[width=0.32\columnwidth]{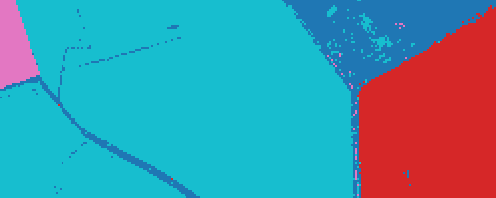}&
    \includegraphics[width=0.32\columnwidth]{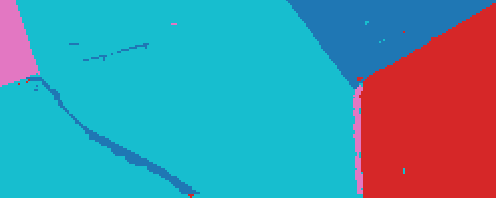}\\
    (d) & (e) & (f) \\
    \includegraphics[width=0.32\columnwidth]{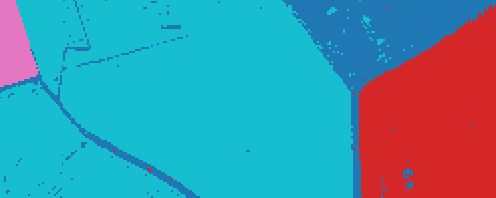}&
    \includegraphics[width=0.32\columnwidth]{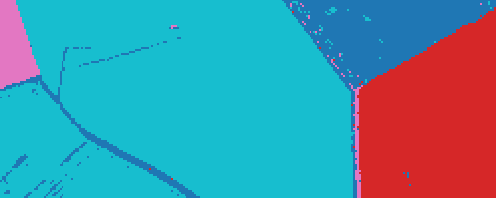}&
    \includegraphics[width=0.32\columnwidth]{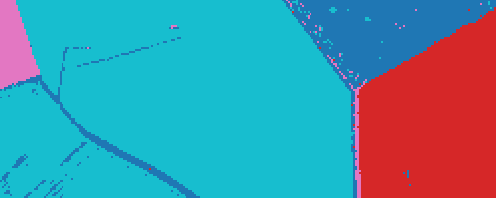}\\
    (g) & (h) & (i) \\
    \includegraphics[width=0.32\columnwidth]{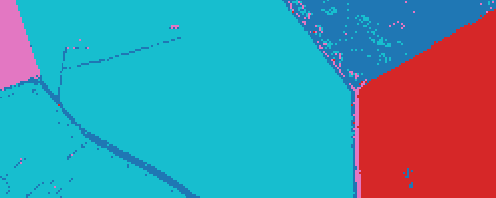}&
    \includegraphics[width=0.32\columnwidth]{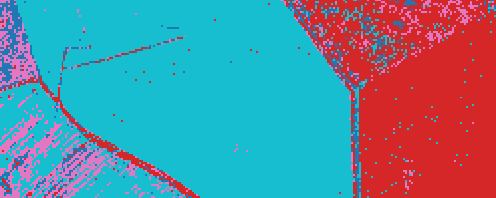}& \\
    (j) & (k) &  \\
  \end{tabular}
  \caption{Synthetic data, classification maps: (a) groundtruth, (b) Cofact-Q, (c) Cofact-CE, (d) MLR, (e) RF, (f) ResNet, (g) SSFPCA, (h) \textsc{fc}-SUnSAL+MLR, (i) \textsc{csr}-SUnSAL+MLR, (j) LC-KSVD, (k) D-KSVD.\label{fig:semisynth-classif}}
\end{figure}

\noindent{\textbf{Performance evaluation --}} Quantitative results obtained on the synthetic dataset are reported in Table~\ref{tab:res-synth} and are visually depicted in Figures \ref{fig:semisynth-classif} and  \ref{fig:semisynth-abund} for the classification and abundance maps, respectively. Metrics and their standard deviation have been computed over $20$ trials. For each trial, a Gaussian white noise is added the observed image such that $\text{SNR} = 30$ dB. From these results, the proposed method appears to be competitive with the compared state-of-the-art methods.
In term of classification results, even though the spatial regularization is very weak in this setting, the cofactorization methods are as good as the RF classifier, which is very satisfying since this latter classifier is one of the most prominent one to deal with HS images~\cite{Stoian2019}. \revBis{The ResNet algorithm shows similar accuracy in term of F1-score but seems to perform slightly better in term of kappa.} However, classification results of \textsc{fc}-SUnSAL and \textsc{csr}-SUnSAL show that a classifier using abundance vectors can already perform well on this toy example where classes are linearly separable. \rev{Similarly, the SSFPCA+SVM methods appears to give interesting results with this synthetic dataset.} \revBis{The MLR using directly the observations appears to be a little less accurate, which may result from the difficulty inherent to high-dimensional inputs.} As for LC-KSVD, it performs slightly worse regarding the F1-mean score whereas results of D-KSVD are clearly the worst.
In term of unmixing performance, \textsc{fc}-SUnSAL, \textsc{csr}-SUnSAL, Cofact-Q and Cofact-CE obtain very similar REs. Note however this metrics only evaluates the quality of the reconstructed data. However, the RMSE is lower with the cofactorization methods and the abundance estimations provided by \textsc{fc}-SUnSAL and \textsc{csr}-SUnSAL significantly degrade.
Even if it is not possible to produce a quantitative evaluation of the representation learnt by D-KSVD and LC-KSVD, REs tends to show that D-KSVD successfully estimated a representation of the data (without being easily interpretable) whereas LC-KSVD seems to focus mostly on the discriminative power of the representation at the price of an inaccurate representation. Moreover, the results produced by LC-KSVD have been obtained by increasing the dimension of the representation $R$ to $40$ while the results obtained by the other methods have been obtained  for $R=15$ to get good classification performances.
The rather poor performance obtained by these two dictionary learning methods, when compared to the proposed cofactorization model, can be explained by the lack of flexibility of the corresponding models which try to recover a descriptive and discriminative representation simultaneously. On the contrary, some flexibility is offered by the clustering step included in the proposed method.
Finally,  comparison in term of processing times shows that D-KSVD, LC-KSVD and the proposed cofactorization methods are significantly slower, which is expected since these methods conducts representation learning and classification jointly. Nonetheless, the cofactorization methods appears faster than D-KSVD and LC-KSVD. It should be also noted that it is necessary to tune manually the number of iterations when using the two latter methods. Conversely, standard convergence criterion can be implemented for the proposed optimization-based methods.

\begin{figure}[!ht]
    \centering
    \begin{tabular}{@{}c@{~}c}
        \includegraphics[width=0.48\columnwidth]{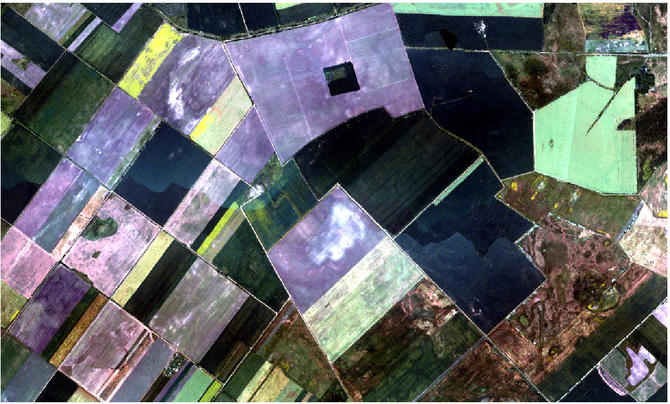} &
        \includegraphics[width=0.48\columnwidth]{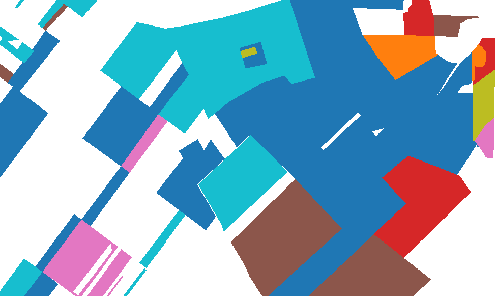}\\
        (a) & (b)
    \end{tabular}
    \caption{AISA dataset: {{(a)}} colored composition of the hyperspectral image $\mathbf{Y}$, {{(b)}} ground-truth [arable land: dark blue, forest: orange, grassland: red, fallowland: brown, leguminosae: pink, reed: green, row crops: light blue].\label{fig:aisa}}
\end{figure}
\begin{table}[!ht]
  \centering
  \caption{AISA data: information about classes.\label{tab:aisa}}
\resizebox{\columnwidth}{!}{%
  \begin{tabular}[b]{lcl}
    \toprule
    Class & Nb. of samples & Subclasses\\
    \midrule
    \multirow{2}{*}{Arable land} & \multirow{2}{*}{177,350} & millet, rape, winter\\
     & &  barley, winter wheat, oat\\
    Forest & 9,274 & forest\\
    Grassland & 25,399 & meadow, pasture\\
    \multirow{2}{*}{Green fallowland} & \multirow{2}{*}{44,370} & fallow treated last year,\\
     & & fallow with shrubs  \\
    Leguminosae & 17,628 & leguminosae\\
    Reed & 4,776 & reed\\
    Row crops & 79,737 & maize, sunflowers\\
    \bottomrule
  \end{tabular}%
}
\end{table}

\subsection{Real hyperspectral image}
\label{sec:real-exp}

\begin{table*}
  \centering
  \caption{AISA data: unmixing and classification results.\label{tab:res-aisa}}
  \begin{tabular}[b]{lcccc}\toprule
    Model & F1-mean & Kappa & RE & Time (s) \\
    \midrule
    Cofact-Q      & $0.503$ ($\pm 4.7\times 10^{-2}$) & $0.652$ ($\pm 2.5\times 10^{-2}$) & $0.310$ ($\pm 1.6\times 10^{-4}$)   & $7303$ ($\pm 139$) \\
    Cofact-CE     & $0.697$ ($\pm 4.5\times 10^{-2}$) & $0.759$ ($\pm 3.5\times 10^{-2}$) & $0.310$ ($\pm 1.4\times 10^{-4}$)   & $4382$ ($\pm 257$) \\
    MLR           & $0.497$ ($\pm 7.3\times 10^{-2}$) & $0.482$ ($\pm 7.7\times 10^{-2}$) & N$\backslash$A                      & $2060$ ($\pm 83$)     \\
    RF            & $0.711$ ($\pm 1.4\times 10^{-2}$) & $0.835$ ($\pm 1.2\times 10^{-2}$) & N$\backslash$A                      & $41$ ($\pm 1$)     \\
    ResNet        & $0.880$ ($\pm 2.3\times 10^{-2}$) & $0.932$ ($\pm 1.3\times 10^{-2}$) & N$\backslash$A                      & $7576$ ($\pm 555$)$^{\star}$     \\
    SSFPCA+SVM    & $0.425$ ($\pm 1.5\times 10^{-2}$) & $0.466$ ($\pm 1.9\times 10^{-2}$) & N$\backslash$A                      & $398$ ($\pm 12$)   \\
    \textsc{fc}-SUnSAL+MLR     & $0.344$ ($\pm 3.1\times 10^{-2}$) & $0.433$ ($\pm 3.8\times 10^{-2}$) & $0.298$ ($\pm 1.9\times 10^{-3}$)   & $512$ ($\pm 96$)   \\
    \textsc{csr}-SUnSAL+MLR    & $0.535$ ($\pm 5.0\times 10^{-2}$) & $0.618$ ($\pm 8.0\times 10^{-2}$) & $0.304$ ($\pm 2.0\times 10^{-5}$)   & $529$ ($\pm 61$)   \\
    D-KSVD        & $0.224$ ($\pm 2.1\times 10^{-2}$) & $0.406$ ($\pm 9.9\times 10^{-2}$) & $0.303$ ($\pm 7.6\times 10^{-6}$)   & $10475$ ($\pm 129$) \\
    LC-KSVD       & $0.350$ ($\pm 3.2\times 10^{-2}$) & $0.594$ ($\pm 3.0\times 10^{-2}$) & $0.303$ ($\pm 4.0\times 10^{-6}$)   & $3780$ ($\pm 320$) \\
    \bottomrule
    \multicolumn{5}{l}{\footnotesize $^{*}$Based on a GPU implementation run on a computer cluster.} \\
  \end{tabular}
\end{table*}

\noindent{\textbf{Description of the dataset --}} The Aisa dataset was acquired by the AISA Eagle sensor during a flight campaign over Heves, Hungary. It contains $L=252$ bands ranging from $395$ to $975$nm. A set of $C=7$ classes have been defined for a total of $358,534$ referenced pixels, according to the class-wise repartition given in Table~\ref{tab:aisa}. To split the full dataset into two test and train subsets, special care has been taken to ensure that training samples are picked out from distinct areas than test samples. The polygons of the reference map are split in smaller polygons on a regular grid pattern and then 50\% of the polygons are taken randomly for training and the remaining 50\% for testing (see \cite{Lagrange2017} for a similar procedure). Figure~\ref{fig:aisa} shows a colored composition of the image and the classification ground-truth. Several reasons justify the choice of this particular dataset. First, it is very challenging both in term of classification and unmixing mostly because the spectral signatures of the classes are very similar, leading in particular to very correlated endmember spectra in $\mathbf{W}$. Secondly, the ground-truth associated to this image is composed of two levels of classification. Thus, an additional ground-truth is available where the $7$ considered classes have been subdivided into $14$ classes also detailed in Table~\ref{tab:aisa}. These subclasses could be compared to the clustering outputs obtained by the proposed cofactorization method, e.g., to verify either the clusters are consistent with the underlying subclasses.\\

\begin{figure}[!ht]
  \centering
    \begin{tikzpicture}
      \begin{axis}[width=0.9\columnwidth,ymin=0.,grid,axis x line=left,axis y line=left,xlabel={$\lambda$ (nm)},ylabel={Reflectance},ylabel style={align=center},font=\footnotesize]
        \addplot+[thick] table[x = wav,y index=0] {Fig/endm-aisa.txt};
        \addplot+[thick] table[x = wav,y index=1] {Fig/endm-aisa.txt};
        \addplot+[thick] table[x = wav,y index=2] {Fig/endm-aisa.txt};
        \addplot+[thick] table[x = wav,y index=3] {Fig/endm-aisa.txt};
        \addplot+[thick] table[x = wav,y index=4] {Fig/endm-aisa.txt};
        \addplot+[thick] table[x = wav,y index=5] {Fig/endm-aisa.txt};
        \addplot+[thick] table[x = wav,y index=6] {Fig/endm-aisa.txt};
        \addplot+[thick] table[x = wav,y index=7] {Fig/endm-aisa.txt};
        \addplot+[thick] table[x = wav,y index=8] {Fig/endm-aisa.txt};
        \addplot+[thick] table[x = wav,y index=9] {Fig/endm-aisa.txt};
        \addplot+[thick] table[x = wav,y index=10] {Fig/endm-aisa.txt};
        \addplot+[thick] table[x = wav,y index=11] {Fig/endm-aisa.txt};
        \addplot+[thick] table[x = wav,y index=12] {Fig/endm-aisa.txt};
      \end{axis};
    \end{tikzpicture}
  \caption{AISA data: spectra used as the dictionary $\mathbf{W}$ identified by the self-dictionary method.\label{fig:aisa-spectra}}
\end{figure}
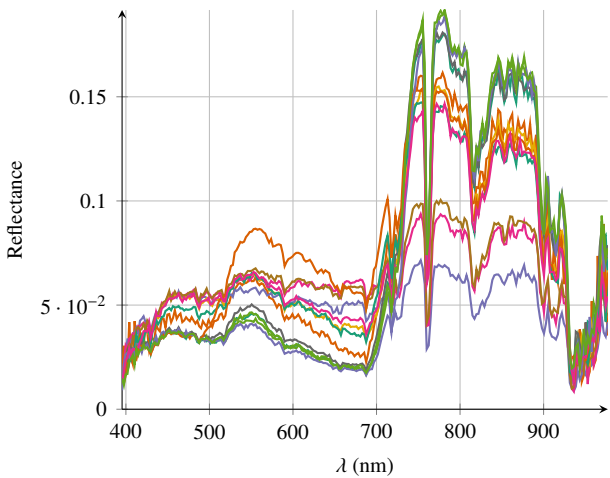

\noindent{\textbf{Compared methods --}} The proposed algorithm is compared to the same methods introduced above. However, note that the D-KSVD method has experienced some difficulties to scale with the size of this new dataset, which is significantly bigger. Thus to obtain results in a decent amount of time, the algorithm has been interrupted prematurely, i.e., before convergence. \rev{Similarly, SVM classifier encounters the same difficulty for the training step and the SVM was finally trained using a subset of the training set (1 over 10 samples).} For the proposed cofactorization method, regularization parameters have been set to  $\tilde{\lambda}_0 = \tilde{\lambda}_1 = \tilde{\lambda}_2 = \tilde{\lambda}_{c} = 1.$ and $\tilde{\lambda}_h =\tilde{\lambda}_{{q}} = 0.01$ and the number of clusters to $K=30$. The initialization step described in Section~\ref{sec:hs-impl} has been performed and the resulting dictionary $\mathbf{W}$ is depicted in Figure~\ref{fig:aisa-spectra} ($R=13$). The same dictionary has been used for the compared unmixing methods.\\

\pgfplotstableset{
    create on use/wav/.style={
        create col/copy column from table={Fig/Res-aisa/wavelength.txt}{Wavelength}
    }
}

\begin{figure}[!ht]
  \centering
  \begin{tabular}{@{}c@{~}c@{~}c}
    \includegraphics[width=0.32\columnwidth]{Fig/aisa_gt_cat.png}&
    \includegraphics[width=0.32\columnwidth]{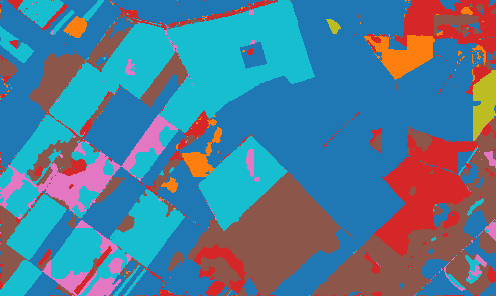}&
    \includegraphics[width=0.32\columnwidth]{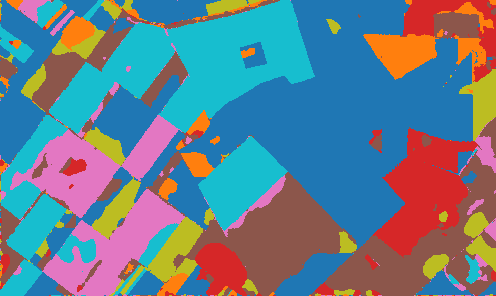}\\
    (a) & (b) & (c) \\
    \includegraphics[width=0.32\columnwidth]{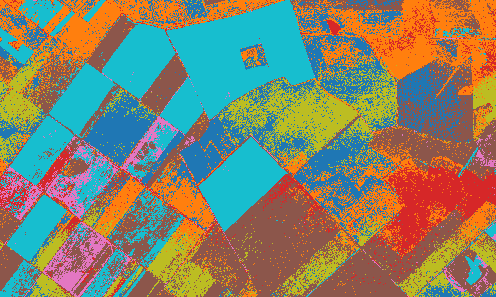}&
    \includegraphics[width=0.32\columnwidth]{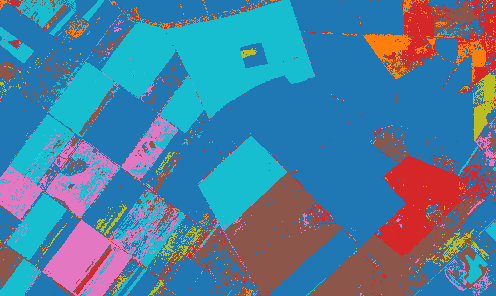}&
    \includegraphics[width=0.32\columnwidth]{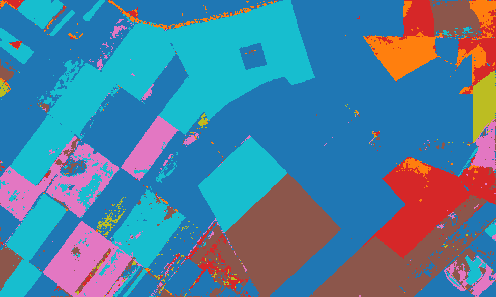}\\
    (d) & (e) & (f) \\
    \includegraphics[width=0.32\columnwidth]{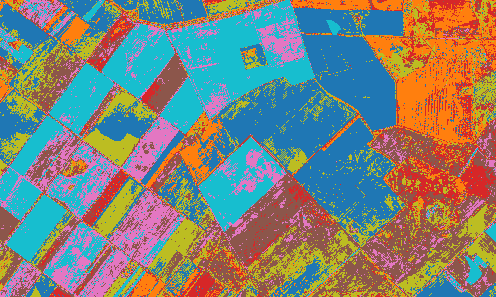}&
    \includegraphics[width=0.32\columnwidth]{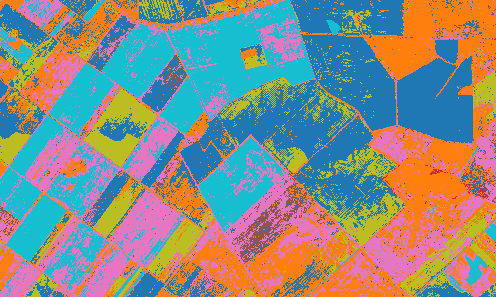}&
    \includegraphics[width=0.32\columnwidth]{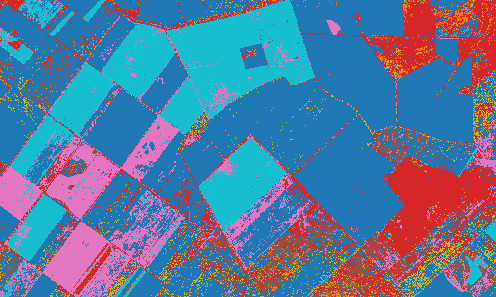}\\
    (g) & (h) & (i) \\
    \includegraphics[width=0.32\columnwidth]{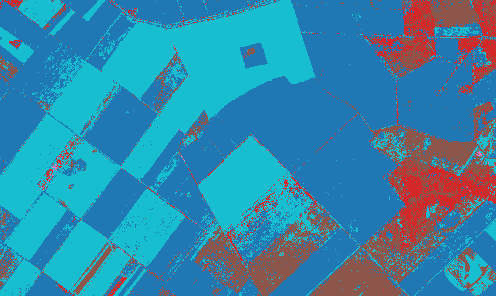}&
    \includegraphics[width=0.32\columnwidth]{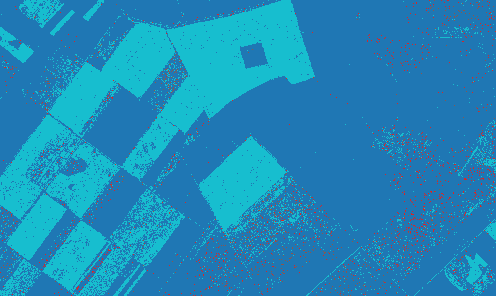}& \\
    (j) & (k) & \\
  \end{tabular}
  \caption{AISA image, classification maps: (a) groundtruth, (b) Cofact-Q, (c) Cofact-CE, (d) MLR, (e) RF, (f) ResNet, (g) SSFPCA, (h) \textsc{fc}-SUnSAL+MLR, (i) \textsc{csr}-SUnSAL+MLR, (j) LC-KSVD, (k) D-KSVD.\label{fig:aisa-classif}}
\end{figure}

\begin{figure*}
  \centering
  \begin{tabular}{@{}cccccc}
    \includegraphics[width=0.3\columnwidth]{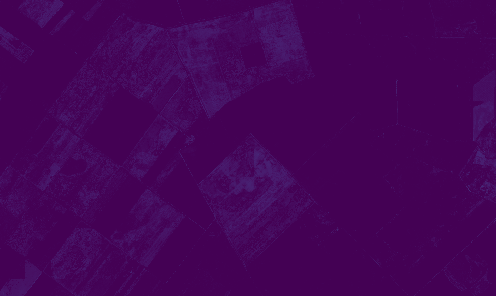}&
    \includegraphics[width=0.3\columnwidth]{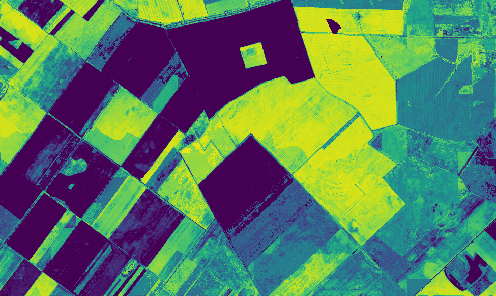}&
    \includegraphics[width=0.3\columnwidth]{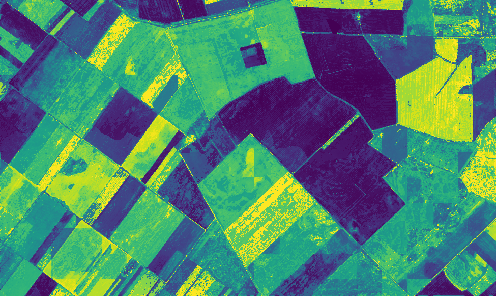}&
    \includegraphics[width=0.3\columnwidth]{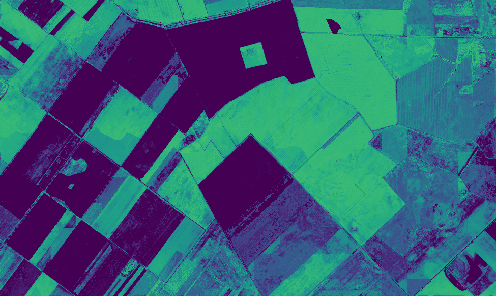}&
    \includegraphics[width=0.3\columnwidth]{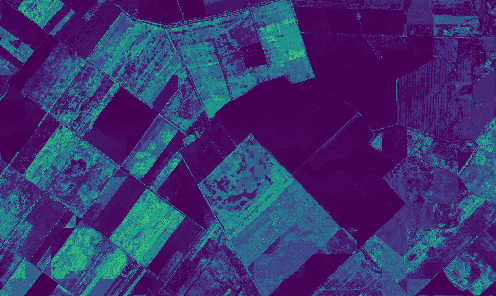}&
    \includegraphics[width=0.3\columnwidth]{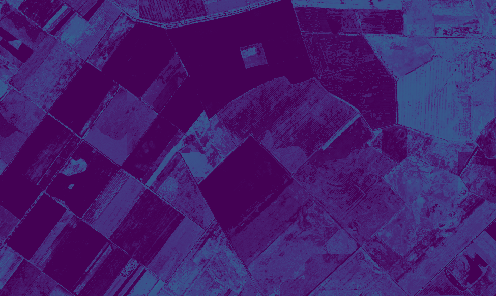}\\
    \includegraphics[width=0.3\columnwidth]{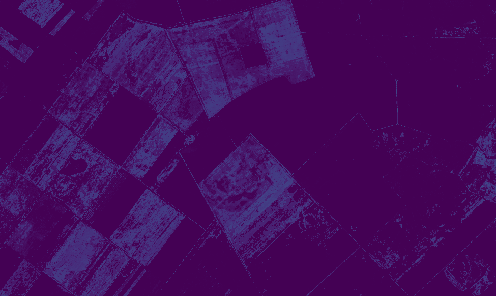}&
    \includegraphics[width=0.3\columnwidth]{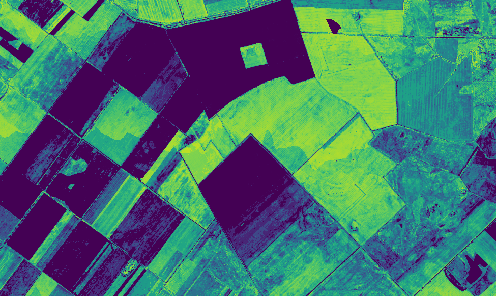}&
    \includegraphics[width=0.3\columnwidth]{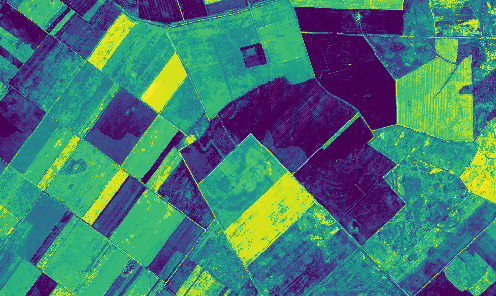}&
    \includegraphics[width=0.3\columnwidth]{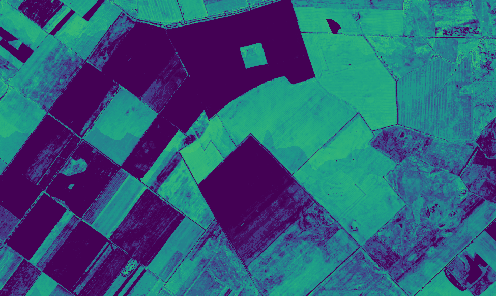}&
    \includegraphics[width=0.3\columnwidth]{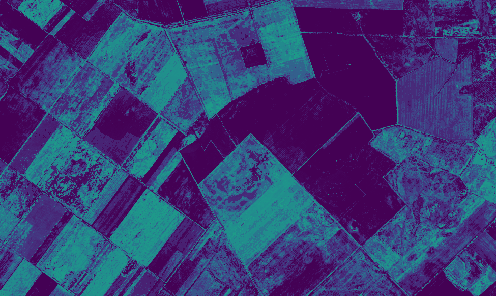}&
    \includegraphics[width=0.3\columnwidth]{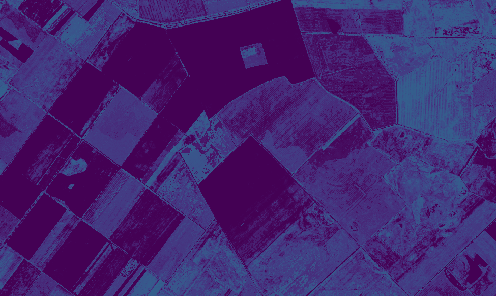}\\
    \includegraphics[width=0.3\columnwidth]{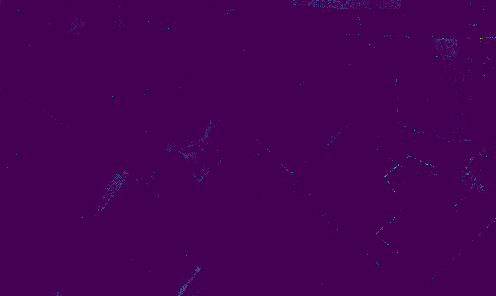}&
    \includegraphics[width=0.3\columnwidth]{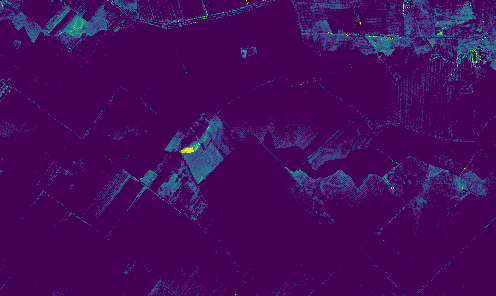}&
    \includegraphics[width=0.3\columnwidth]{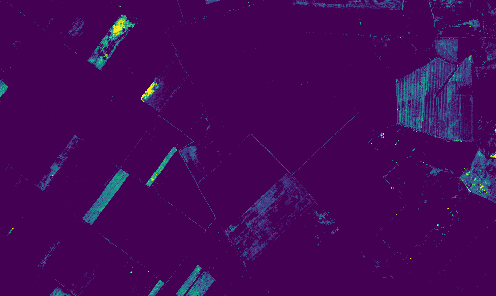}&
    \includegraphics[width=0.3\columnwidth]{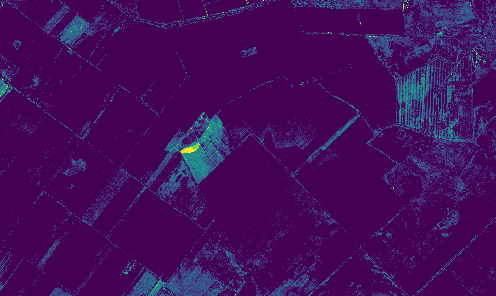}&
    \includegraphics[width=0.3\columnwidth]{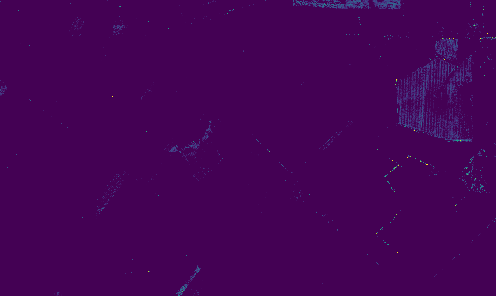}&
    \includegraphics[width=0.3\columnwidth]{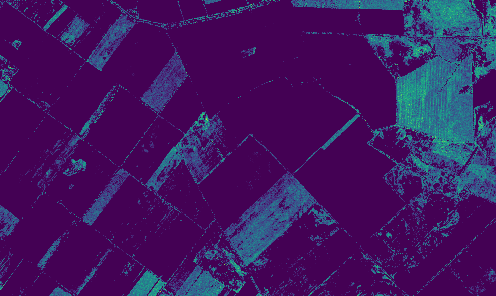}\\
    \includegraphics[width=0.3\columnwidth]{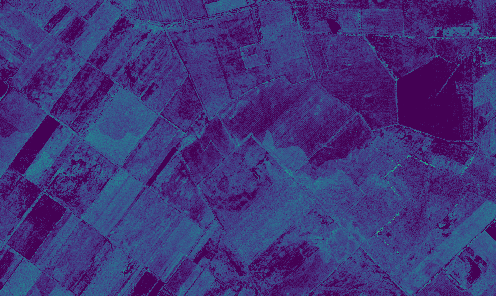}&
    \includegraphics[width=0.3\columnwidth]{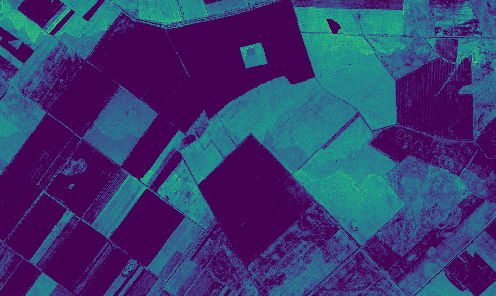}&
    \includegraphics[width=0.3\columnwidth]{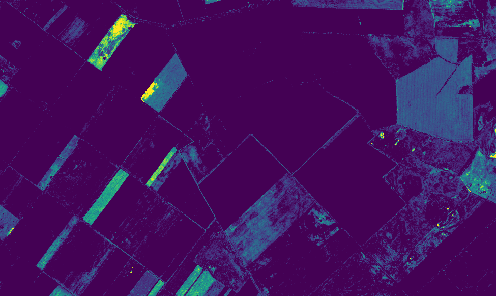}&
    \includegraphics[width=0.3\columnwidth]{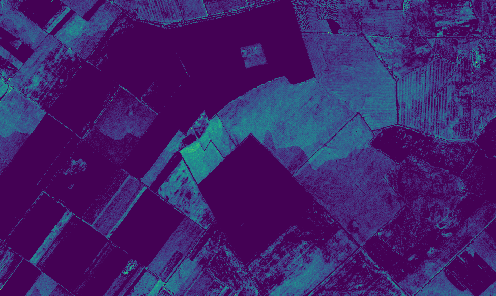}&
    \includegraphics[width=0.3\columnwidth]{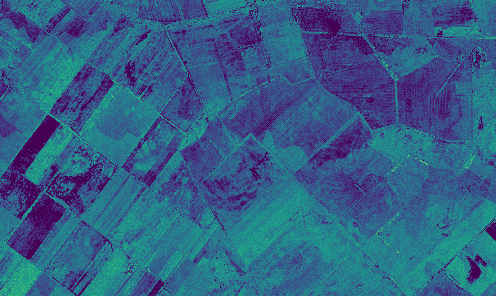}&
    \includegraphics[width=0.3\columnwidth]{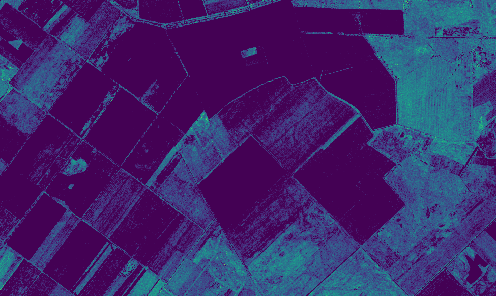}\\
  \end{tabular}
  \caption{AISA dataset, abundances map for the $6$ components: ($1$st row) Cofact-Q, ($2$nd row) Cofact-CE, ($3$rd row) \textsc{fc}-SUnSAL and ($4$th row) \textsc{csr}-SUnSAL.\label{fig:aisa-abund}}
\end{figure*}

\noindent{\textbf{Performance evaluation --}} All quantitative results are presented in Table~\ref{tab:res-aisa}. Metrics and their standard deviation have been computed over $5$ trials. RMSE metrics have been removed since no groundtruth is available to assess the quality of the estimated abundance maps. RE is thus the only used figure-of-merit to assess the quality of the representation learning. Note however, as previously explained, RE does not directly evaluate the correctness of the abundance maps. In the present case, REs appear to be very similar for all algorithms. Contrary to the previous dataset, this is also the case for LC-KSVD, which can be explained by the fact that spectra are similar in the whole image and it is thus quite easy to get a very low RE with any estimated dictionary. This is the reason why qualitative evaluation remains interesting. Figure~\ref{fig:aisa-abund} shows a subset of the estimated abundance maps. It is difficult to draw any incontestable conclusion but it is clear that, despite similar REs, significantly different result are obtained for each method. This behavior is strengthened by the very high correlation between the endmembers in this dataset, which may lead to probable mismatch between endmember spectra. Nevertheless the Cofact methods seems to give slightly more consistent results. Indeed, edges in the abundance maps appear to be more consistent with boundaries observed in the hyperspectral image. Additionally, for the compared methods, some abundance maps seem to be influenced by the presence of two flight lines in the image. This phenomenon clearly appears in the abundance maps recovered by \textsc{fc}-SUnSAL ($3$rd row).

Concerning classification results, the results reported in Table~\ref{tab:res-aisa} show that the classification maps recovered by the Cofact-CE is very closed to the one obtained by RF, \rev{whereas SSFPCA+SVM fails to provide reasonable results}. \revBis{As for the ResNet method, it clearly outperforms all the other methods.} \revTer{The better performance could be explained by the fact the neural network used convolutional layers which extract spatial context information. On the contrary, the other methods rely on pixelwise inputs with, at best, a spatial regularization which only promotes local regularity without benefiting from a richer description of the spatial context.} Figure~\ref{fig:aisa-classif} shows in particular that the cofactorization methods encounter some trouble distinguishing very similar classes, for example \emph{grassland} (red) from \emph{fallowland} (brown). Nevertheless, the obtained classification appears to be consistent and it seems reasonable to expect a lesser degradation of the classification results when considering less correlated spectral signatures. This confusion explains the less convincing results of the proposed method with quadratic loss. \revBis{Besides, it is important to keep in mind that the objective of this work is not to propose the most efficient classification method but rather to propose a method that can give results of similar quality than some state-of-the-art methods, with the benefit of providing additional insights thanks to the joint representation learning.} The results also show that the proposed method is beneficial to the classification since \textsc{fc}-SUnSAL+MLR, \textsc{csr}-SUnSAL+MLR\revBis{, MLR} and Cofact-CE use the same classifier and the latter performs clearly better. The comparison between the representation learning-based algorithms is clear and the both Cofact methods perform better than LC-KSVD and D-KSVD.

In term of processing time,  LC-KSVD, D-KSVD and the Cofact methods are clearly more time consuming. Nevertheless, all those methods provide more outputs than the other methods. The comparison between these methods seems to give an advantage for  LC-KSVD. However, it should be noted that it is very difficult to monitor the convergence of LC-KSVD and D-KSVD since the value of the objective function over the iteration is not monotonic. The proposed algorithms and their implementations thus give a practical advantage since they do not need to be applied with different numbers of iterations to ensure good results.

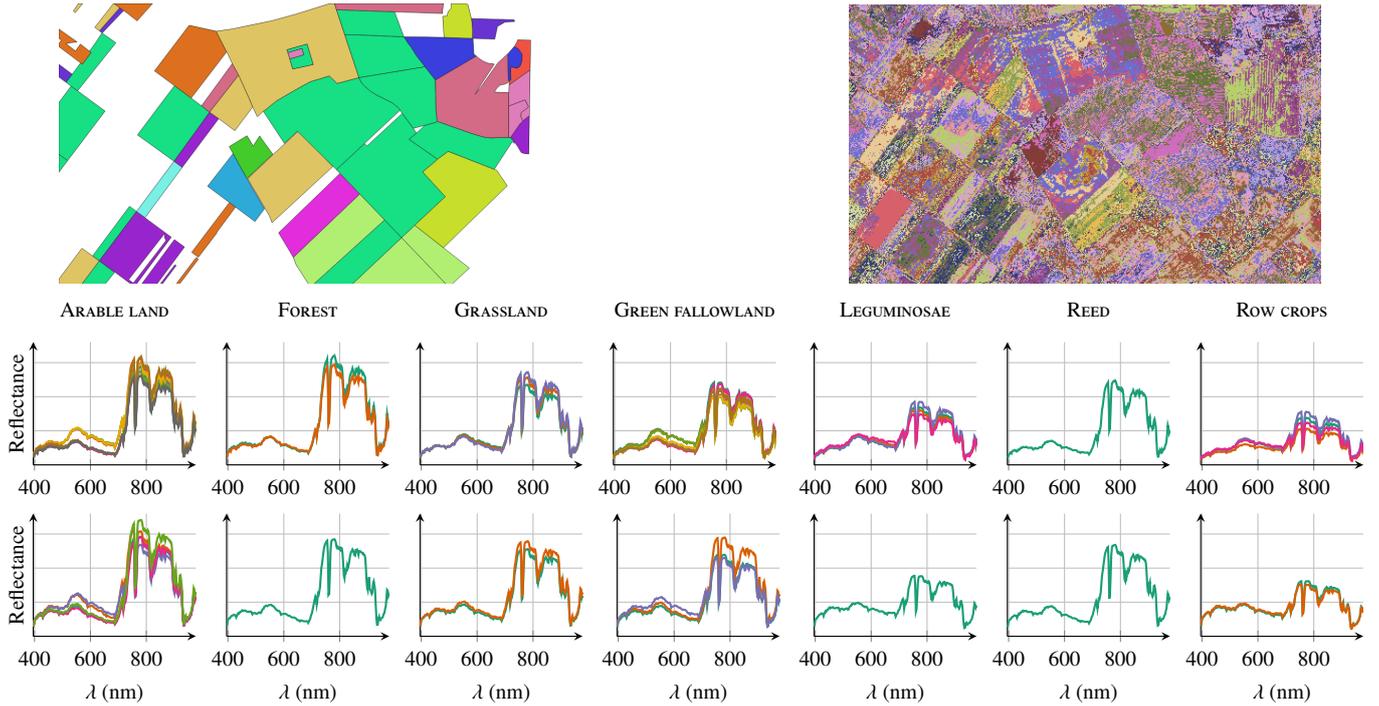
\begin{figure*}
  \centering
  \begin{tabular}{@{}c@{~}c@{~}c@{~}c@{~}c@{~}c@{~}c}
    \multicolumn{3}{c}{\includegraphics[width=0.7\columnwidth]{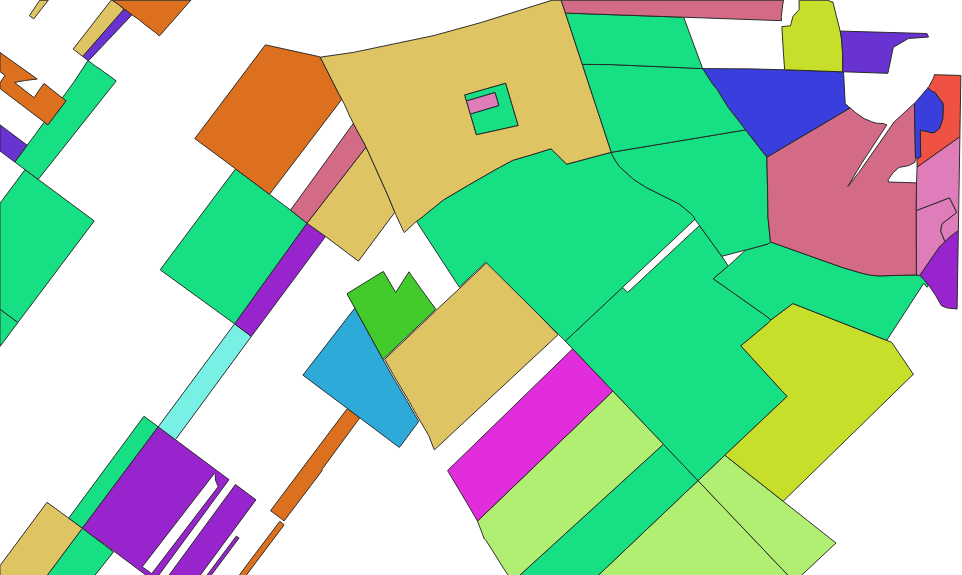}} & &
    \multicolumn{3}{c}{\includegraphics[width=0.7\columnwidth]{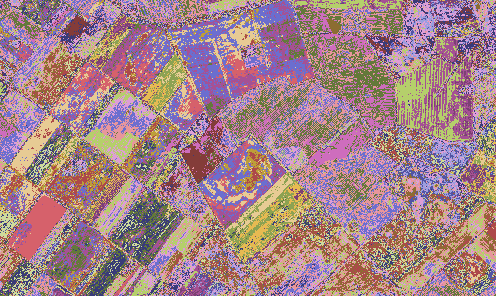}} \\
    \begin{tikzpicture}
      \begin{axis}[width=0.42\columnwidth,ymin=0.,ymax=0.18,ymajorticks=false,grid,axis x line=left,axis y line=left,ylabel={Reflectance},ylabel style={align=center},font=\footnotesize,title=\sc{Arable land}]
        \addplot+[thick] table[x = wav,y index=0] {Fig/Res-aisa/aisa_clustOfClass0.txt};
        \addplot+[thick] table[x = wav,y index=1] {Fig/Res-aisa/aisa_clustOfClass0.txt};
        \addplot+[thick] table[x = wav,y index=2] {Fig/Res-aisa/aisa_clustOfClass0.txt};
        \addplot+[thick] table[x = wav,y index=3] {Fig/Res-aisa/aisa_clustOfClass0.txt};
        \addplot+[thick] table[x = wav,y index=4] {Fig/Res-aisa/aisa_clustOfClass0.txt};
        \addplot+[thick] table[x = wav,y index=5] {Fig/Res-aisa/aisa_clustOfClass0.txt};
        \addplot+[thick] table[x = wav,y index=6] {Fig/Res-aisa/aisa_clustOfClass0.txt};
        \addplot+[thick] table[x = wav,y index=7] {Fig/Res-aisa/aisa_clustOfClass0.txt};
      \end{axis};
    \end{tikzpicture}&
    \begin{tikzpicture}
      \begin{axis}[width=0.42\columnwidth,ymin=0.,ymax=0.18,ymajorticks=false,grid,axis x line=left,axis y line=left,font=\footnotesize,title=\sc{Forest}]
        \addplot+[thick] table[x = wav,y index=0] {Fig/Res-aisa/aisa_clustOfClass1.txt};
        \addplot+[thick] table[x = wav,y index=1] {Fig/Res-aisa/aisa_clustOfClass1.txt};
      \end{axis};
    \end{tikzpicture}&
    \begin{tikzpicture}
      \begin{axis}[width=0.42\columnwidth,ymin=0.,ymax=0.18,ymajorticks=false,grid,axis x line=left,axis y line=left,font=\footnotesize,title=\sc{Grassland}]
        \addplot+[thick] table[x = wav,y index=0] {Fig/Res-aisa/aisa_clustOfClass2.txt};
        \addplot+[thick] table[x = wav,y index=1] {Fig/Res-aisa/aisa_clustOfClass2.txt};
        \addplot+[thick] table[x = wav,y index=2] {Fig/Res-aisa/aisa_clustOfClass2.txt};
      \end{axis};
    \end{tikzpicture}&
    \begin{tikzpicture}
      \begin{axis}[width=0.42\columnwidth,ymin=0.,ymax=0.18,ymajorticks=false,grid,axis x line=left,axis y line=left,font=\footnotesize,title=\sc{Green fallowland}]
        \addplot+[thick] table[x = wav,y index=0] {Fig/Res-aisa/aisa_clustOfClass3.txt};
        \addplot+[thick] table[x = wav,y index=1] {Fig/Res-aisa/aisa_clustOfClass3.txt};
        \addplot+[thick] table[x = wav,y index=2] {Fig/Res-aisa/aisa_clustOfClass3.txt};
        \addplot+[thick] table[x = wav,y index=3] {Fig/Res-aisa/aisa_clustOfClass3.txt};
        \addplot+[thick] table[x = wav,y index=4] {Fig/Res-aisa/aisa_clustOfClass3.txt};
        \addplot+[thick] table[x = wav,y index=5] {Fig/Res-aisa/aisa_clustOfClass3.txt};
        \addplot+[thick] table[x = wav,y index=6] {Fig/Res-aisa/aisa_clustOfClass3.txt};
      \end{axis};
    \end{tikzpicture}&
    \begin{tikzpicture}
      \begin{axis}[width=0.42\columnwidth,ymin=0.,ymax=0.18,ymajorticks=false,grid,axis x line=left,axis y line=left,font=\footnotesize,title=\sc{Leguminosae}]
        \addplot+[thick] table[x = wav,y index=0] {Fig/Res-aisa/aisa_clustOfClass4.txt};
        \addplot+[thick] table[x = wav,y index=1] {Fig/Res-aisa/aisa_clustOfClass4.txt};
        \addplot+[thick] table[x = wav,y index=2] {Fig/Res-aisa/aisa_clustOfClass4.txt};
        \addplot+[thick] table[x = wav,y index=3] {Fig/Res-aisa/aisa_clustOfClass4.txt};
      \end{axis};
    \end{tikzpicture}&
    \begin{tikzpicture}
      \begin{axis}[width=0.42\columnwidth,ymin=0.,ymax=0.18,ymajorticks=false,grid,axis x line=left,axis y line=left,font=\footnotesize,title=\sc{Reed}]
        \addplot+[thick] table[x = wav,y index=0] {Fig/Res-aisa/aisa_clustOfClass5.txt};
      \end{axis};
    \end{tikzpicture}&
    \begin{tikzpicture}
      \begin{axis}[width=0.42\columnwidth,ymin=0.,ymax=0.18,ymajorticks=false,grid,axis x line=left,axis y line=left,font=\footnotesize,title=\sc{Row crops}]
        \addplot+[thick] table[x = wav,y index=0] {Fig/Res-aisa/aisa_clustOfClass6.txt};
        \addplot+[thick] table[x = wav,y index=1] {Fig/Res-aisa/aisa_clustOfClass6.txt};
        \addplot+[thick] table[x = wav,y index=2] {Fig/Res-aisa/aisa_clustOfClass6.txt};
        \addplot+[thick] table[x = wav,y index=3] {Fig/Res-aisa/aisa_clustOfClass6.txt};
      \end{axis};
    \end{tikzpicture}\\
    \begin{tikzpicture}
      \begin{axis}[width=0.42\columnwidth,ymin=0.,ymax=0.18,ymajorticks=false,grid,axis x line=left,axis y line=left,xlabel={$\lambda$ (nm)},ylabel={Reflectance},ylabel style={align=center},font=\footnotesize]
        \addplot+[thick] table[x = wav,y index=7] {Fig/subcat_mean.txt};
        \addplot+[thick] table[x = wav,y index=9] {Fig/subcat_mean.txt};
        \addplot+[thick] table[x = wav,y index=13] {Fig/subcat_mean.txt};
        \addplot+[thick] table[x = wav,y index=14] {Fig/subcat_mean.txt};
        \addplot+[thick] table[x = wav,y index=15] {Fig/subcat_mean.txt};
      \end{axis};
    \end{tikzpicture}&
    \begin{tikzpicture}
      \begin{axis}[width=0.42\columnwidth,ymin=0.,ymax=0.18,ymajorticks=false,grid,axis x line=left,axis y line=left,xlabel={$\lambda$ (nm)},font=\footnotesize]
        \addplot+[thick] table[x = wav,y index=1] {Fig/subcat_mean.txt};
      \end{axis};
    \end{tikzpicture}&
    \begin{tikzpicture}
      \begin{axis}[width=0.42\columnwidth,ymin=0.,ymax=0.18,ymajorticks=false,grid,axis x line=left,axis y line=left,xlabel={$\lambda$ (nm)},font=\footnotesize]
        \addplot+[thick] table[x = wav,y index=6] {Fig/subcat_mean.txt};
        \addplot+[thick] table[x = wav,y index=8] {Fig/subcat_mean.txt};
      \end{axis};
    \end{tikzpicture}&
    \begin{tikzpicture}
      \begin{axis}[width=0.42\columnwidth,ymin=0.,ymax=0.18,ymajorticks=false,grid,axis x line=left,axis y line=left,xlabel={$\lambda$ (nm)},font=\footnotesize]
        \addplot+[thick] table[x = wav,y index=2] {Fig/subcat_mean.txt};
        \addplot+[thick] table[x = wav,y index=3] {Fig/subcat_mean.txt};
        \addplot+[thick] table[x = wav,y index=4] {Fig/subcat_mean.txt};
      \end{axis};
    \end{tikzpicture}&
    \begin{tikzpicture}
      \begin{axis}[width=0.42\columnwidth,ymin=0.,ymax=0.18,ymajorticks=false,grid,axis x line=left,axis y line=left,xlabel={$\lambda$ (nm)},font=\footnotesize]
        \addplot+[thick] table[x = wav,y index=0] {Fig/subcat_mean.txt};
      \end{axis};
    \end{tikzpicture}&
    \begin{tikzpicture}
      \begin{axis}[width=0.42\columnwidth,ymin=0.,ymax=0.18,ymajorticks=false,grid,axis x line=left,axis y line=left,xlabel={$\lambda$ (nm)},font=\footnotesize]
        \addplot+[thick] table[x = wav,y index=10] {Fig/subcat_mean.txt};
      \end{axis};
    \end{tikzpicture}&
    \begin{tikzpicture}
      \begin{axis}[width=0.42\columnwidth,ymin=0.,ymax=0.18,ymajorticks=false,grid,axis x line=left,axis y line=left,xlabel={$\lambda$ (nm)},font=\footnotesize]
        \addplot+[thick] table[x = wav,y index=5] {Fig/subcat_mean.txt};
        \addplot+[thick] table[x = wav,y index=11] {Fig/subcat_mean.txt};
      \end{axis};
    \end{tikzpicture}\\
  \end{tabular}
  \caption{AISA data: ($1$st row) Groundtruth map of subclasses and clustering recovered by Cofact-CE, ($2$nd row) for each class, spectral centroids of the clusters recovered by Cofact-CE composing the class, ($3$rd row) for each class, mean spectra of the groundtruth subclasses composing the class.\label{fig:clust-means}}
\end{figure*}

One of very interesting feature of the Cofact method is the possibility of examining the clusters obtained as a byproduct. Given the formulation \eqref{eq:clust-cstr-pb}, the centroids $\mathbf{B}$ estimated by the Cofact method can be interpreted as average behaviors of abundance vectors. Corresponding virtual spectral signatures can be obtained by right-multiplying the dictionary $\mathbf{W}$ by this estimated abundance-like matrix $\mathbf{B}$. The first line in Figure~\ref{fig:clust-means} shows these spectral centroids for each cluster. Accessing this kind of information is precious in term of image interpretation since it offers the possibility of visualizing any class multi-modality. To illustrate, the second line of Figure~\ref{fig:clust-means} shows the mean spectra associated with the subclass groundtruth. Clearly, both lines exhibit strong similarities, with spectral diversity (hence multi-modality) for the $1$st, $3$rd and $4$rd classes. This illustrates the relevance of the clusters recovered by the proposed cofactorization method.

\section{Conclusion and perspectives}
\label{sec:ccl}

This paper proposed a cofactorization model to unify a representation learning task and a classification task. The coding matrices associated with the two factorization problems, which respectively are the low-dimensional representations and the feature vectors, were related thanks to a clustering step. The low-dimensional representation vectors were clustered and the resulting attribution vectors were used as features vectors. These three tasks were jointly formulated as a non-convex non-smooth minimization problem, whose solution was approximated thanks to a PALM algorithm which ensured some convergence guarantees. \rev{The interest of considering a clustering task as a coupling process is threefold. First, it allows the learnt representation to be both descriptive and discriminative to ensure a low reconstruction error and a good separability of the classes, respectively. These two properties are often adversarial and the clustering term offers an additional degree of freedom to accommodate both properties. Secondly, instead of linearly separating the classes in the low-dimensional representation space, the resulting method achieves a non-linear classification relying on the coding vectors. The clustering term acts similarly as the well-known kernel trick since the coding vectors are mapped into a new representation space, the cluster attribution space, where classes are expected to be linearly separable. Finally, the clustering is very interesting to interpret the obtained results. For instance, analyzing the identified cluster centroids allows the end-user to characterize the possible class multi-modality.} This model was instanced in an particular applicative scenario, namely hyperspectral image analysis, to jointly conduct unmixing and classification. It provided convincing results on synthetic and real data both quantitatively and qualitatively. Moreover, byproducts of the estimation appeared to be a relevant added value to interpret the obtained results.

To further improve the developed model, it would be particularly interesting to investigate the best way to learn an appropriate dictionary. For instance, it would be relevant to directly exploit the supervised information to get a better dictionary initialization. Moreover, updating the dictionary when solving the cofactorization problem would be also of interest. \revTer{Another promising future work would consist in replacing the stage of the model dedicated to the classification task by a more advanced classifier. Indeed, when using the cross-entropy loss, this factorization model was interpreted as a single-layer neural network. A natural extension would be to leverage on a deeper architecture, while preserving the benefit of interpretability brought by the hierarchical representation of the data through the learning, clustering and classification steps.} Finally, the genericity of the proposed approach should be assessed through the analysis of data from other applicative contexts where representation learning and classification play central roles, such as medical imaging of various modalities \cite{Chaari2013,Cavalcanti2018}.

\section*{Acknowledgements}
The authors would like to thank Olivier Gouvert and Prof. C\'edric F\'evotte (IRIT, Univ. of Toulouse, CNRS, France) for fruitful discussions regarding this work. They also thank Juan Mario Haut and Pr. Antonio J. Plaza for providing the code associated with the ResNet method \cite{Paoletti2019}. Part of the numerical experiments were conducted on the OSIRIM platform\footnote{\url{http://osirim.irit.fr/site/en}} of IRIT, supported by the CNRS, the FEDER, the Occitanie region and the French government.

\appendix
\section{Technical derivations}

This appendix provides some details regarding the optimization schemes instanced for the proposed cofactorization model with the classification quadratic and cross-entropy losses.

\label{sec:hs-optim}
\subsection{Cofactorization model with quadratic loss function}
Using notations consistent with \eqref{eq:palm-pb}, the smooth coupling term of the quadratic (Q) loss cost can be expressed as\vspace{-0.15cm}
\begin{align*}
g&(\mathbf{H},\mathbf{B},\mathbf{Z},\mathbf{C}_{\mathcal{U}},\mathbf{Q})
    =
   \frac{\lambda_0}{2} \norm{\mathbf{Y} - \mathbf{W}\mathbf{H}}_{\mathrm{F}}^2  \nonumber\\
  & + \frac{\lambda_1}{2} \norm{\mathbf{C}\mathbf{D} - \mathbf{Q}\mathbf{Z}\mathbf{D}}_{\mathrm{F}}^2 + \lambda_{c} \norm{\mathbf{C}}_{\mathrm{vTV}} + \frac{\lambda_2}{2} \norm{\mathbf{H} - \mathbf{B}\mathbf{Z}}_{\mathrm{F}}^2.\vspace{-0.10cm}
\end{align*}
For a practical implementation, one needs to compute the partial gradients of $g(\cdot)$ explicitly and their Lipschitz constants to perform the gradient descent. Regarding the $\mathbf{H}$ and $\mathbf{B}$ variables, these computations are the same for the two models (quadratic and cross-entropy losses) and lead to\vspace{-0.10cm}
\begin{align}
\nabla_\mathbf{H} g(\mathbf{H},\mathbf{B},\mathbf{Z},\mathbf{C}_{\mathcal{U}},\mathbf{Q}) &= \lambda_0 (\mathbf{W}^t\mathbf{W}\mathbf{H} - \mathbf{W}^t\mathbf{Y}) \\ &+ \lambda_2 (\mathbf{H} - \mathbf{B}\mathbf{Z}), \nonumber\\
\nabla_\mathbf{B} g(\mathbf{H},\mathbf{B},\mathbf{Z},\mathbf{C}_{\mathcal{U}},\mathbf{Q}) &= \lambda_2 (\mathbf{B}\mathbf{Z}\mathbf{Z}^t - \mathbf{H}\mathbf{Z}^t),\vspace{-0.10cm}
\end{align}
Regarding the variables $\mathbf{Z}$, $\mathbf{Q}$ and $\mathbf{C}_{\mathcal{U}}$ involved in the classification step with quadratic loss, they writes
\begin{align}
\nabla_\mathbf{Z} g(\mathbf{H},\mathbf{B},\mathbf{Z},\mathbf{C}_{\mathcal{U}},\mathbf{Q}) &= \lambda_2 (\mathbf{B}^T\mathbf{B}\mathbf{Z} - \lambda_1 \mathbf{B}^T\mathbf{H}) \nonumber\\
& + \lambda_1 (\mathbf{Q}^T\mathbf{Q}\mathbf{Z}\mathbf{D}^2 - \mathbf{Q}^T\mathbf{C}\mathbf{D}^2), \nonumber\\
\nabla_\mathbf{Q} g(\mathbf{H},\mathbf{B},\mathbf{Z},\mathbf{C}_{\mathcal{U}},\mathbf{Q}) &= \lambda_1 (\mathbf{Q}\mathbf{Z}\mathbf{D}^2\mathbf{Z}^T - \mathbf{C}\mathbf{D}^2\mathbf{Z}^T), \nonumber\\
\nabla_{\mathbf{C}_{\mathcal{U}}} g(\mathbf{H},\mathbf{B},\mathbf{Z},\mathbf{C}_{\mathcal{U}},\mathbf{Q}) &= \lambda_{c} \nabla_{\mathbf{C}_{\mathcal{U}}} \norm{\mathbf{C}}_{\mathrm{vTV}} \nonumber\\
&+ \lambda_1 (\mathbf{C}_{\mathcal{U}}\mathbf{D}_{\mathcal{U}}^2 - \mathbf{Q} \mathbf{Z}_{\mathcal{U}}\mathbf{D}_{\mathcal{U}}^2).\vspace{-0.10cm}
\end{align}
For sake of brevity, the gradient $\nabla_{\cdot} \norm{\cdot}_{\mathrm{vTV}}$ of the vectorial TV regularization is not explicitly given. Readers are referred to~\cite{Getreuer2012} for further details.

All partial gradients are globally Lipschitz as functions of the corresponding partial variables. After basic matrix derivations, majorizations similar to \eqref{eq:Lip_constant_def} lead to the following Lipschitz constant\vspace{-0.10cm}
\begin{align}
L_\mathbf{H} &= \norm{\lambda_0 \mathbf{W}^T\mathbf{W} +\lambda_2 \mathbf{I}_R}, \nonumber\\
L_\mathbf{B}(\mathbf{Z}) &= \norm{\lambda_2 \mathbf{Z}\mathbf{Z}^T}, \nonumber\\
L_\mathbf{Z}(\mathbf{B},\mathbf{Q}) &= \max_p \norm{\lambda_2 \mathbf{B}^T\mathbf{B} + \lambda_1 d_{p} \mathbf{Q}^T\mathbf{Q}}, \nonumber\\
L_\mathbf{Q}(\mathbf{Z}) &= \norm{\lambda_1 \mathbf{Z}\mathbf{D}^2\mathbf{Z}^T}, \nonumber\\
L_\mathbf{\mathbf{C}_{\mathcal{U}}} &= \lambda_1 \max_p d_{p}^2 + \lambda_{c} \frac{\sqrt{8}\max_p \beta_p}{\epsilon}.
\end{align}

\subsection{Cofactorization model with cross-entropy loss function}
When using cross-entropy as the classification loss function, the coupling term writes
\begin{align}
g&(\mathbf{H},\mathbf{B},\mathbf{Z},\mathbf{C}_{\mathcal{U}},\mathbf{Q}) =
  \frac{\lambda_0}{2} \norm{\mathbf{Y} - \mathbf{W}\mathbf{H}}_{\mathrm{F}}^2 \nonumber\\
  & - \frac{\lambda_1}{2} \sum_{p \in\mathcal{P}} d_{p}^2 \sum_{i \in\mathcal{C}} c_{i,p} \log\left( \sigm(-\mathbf{q}_{i:} \mathbf{z}_p)) \right) \nonumber\\
  & + \frac{\lambda_{{q}}}{2} \norm{\mathbf{Q}}_F^2 + \lambda_{c} \norm{\mathbf{C}}_{\mathrm{vTV}} + \frac{\lambda_2}{2} \norm{\mathbf{H} - \mathbf{B}\mathbf{Z}}_F^2
\end{align}
and the specific partial gradients are
\begin{align}
\nabla_\mathbf{Z} &g(\mathbf{H},\mathbf{B},\mathbf{Z},\mathbf{C}_{\mathcal{U}},\mathbf{Q}) = - \frac{\lambda_1}{2} \mathbf{Q}^T \mathbf{G} \nonumber\\
\nabla_\mathbf{Q} &g(\mathbf{H},\mathbf{B},\mathbf{Z},\mathbf{C}_{\mathcal{U}},\mathbf{Q}) = - \frac{\lambda_1}{2} \mathbf{G} \mathbf{Z}^T + \lambda_{{q}} \mathbf{Q}, \nonumber\\
\nabla_{\mathbf{C}_{\mathcal{U}}} &g(\mathbf{H},\mathbf{B},\mathbf{Z},\mathbf{C}_{\mathcal{U}},\mathbf{Q}) = \lambda_{c} \nabla_{\mathbf{C}_{\mathcal{U}}} \norm{\mathbf{C}_{\mathcal{U}}}_{\mathrm{vTV}} \nonumber\\
&- \frac{\lambda_1}{2} \sum_{p \in\mathcal{P}} d_{p}^2 \sum_{i \in\mathcal{C}} \log\left( \sigm(-\mathbf{q}_{i:} \mathbf{z}_p)) \right)
\end{align}
where $\mathbf{G}$ is a $C\times P$ matrix with elements given by
\begin{equation}
  g_{i,p} = \frac{d_{p}^2 c_{i,p}}{1+\exp(-\mathbf{q}_{i:} \mathbf{z}_p)}.
\end{equation}
It should be noticed that $\mathbf{G}$ depends on $\mathbf{Z}$, $\mathbf{Q}$ and $\mathbf{C}$ and is only introduced here to get compact notations. The following Lipschitz constants can be derived
\begin{align}
L_\mathbf{Z}(\mathbf{B},\mathbf{Q}) &= \lambda_1 \sum_{p\in\mathcal{P}} d_{p}^2 \sum_{i\in\mathcal{C}} c_{i,p} \norm{\mathbf{q}_{j:}}_2^2 + \norm{\lambda_2 \mathbf{B}\mathbf{B}^T}, \nonumber\\
L_\mathbf{Q} &= \lambda_1 \sum_{p\in\mathcal{P}} d_{p}^2 + \lambda_{\text{q}}, \nonumber\\
L_\mathbf{\mathbf{C}_{\mathcal{U}}} &= \lambda_{c} \frac{\sqrt{8}\max_p \beta_p}{\epsilon}.
\end{align}

\subsection{Computing the proximal operators}

For a practical implementation of the PALM algorithm, the proximal operators associated with each $f_j(\cdot)$ in~\eqref{eq:functions_f} need to be computed. It is clear that all these functions are proper lower semi-continuous functions for both models instanced in Section~\ref{sec:hs-global-pb}. The involved indicator functions are defined on convex sets. Thus, their proximal operators can be expressed as projections. The projection on the non-negative quadrant is a simple thresholding of negative values. The projection on the simplices $\mathcal{S}_{\cdot}$ can be conducted as detailed in~\cite{Condat2016}. The case of $f_0(\cdot)$ defined by a nonnegativity constraint complemented by a $\ell_1$-norm sparsity promoting regularization is slightly more complex. It can be handled using a composition of proximal operators. As stated before, the proximal operator associated to the positivity constraint is the projection on the non-negative quadrant. The proximal operator associated with the $\ell_1$-norm penalization is a soft-thresholding, i.e., {$\prox{\norm{\cdot}_1}{t}(x) = \text{sign}(x) (|x|-\frac{1}{t})_+$}~\cite{Jenatton2011}. These two proximal operators satisfy the conditions exhibited in~\cite{Yu2013} required to be allowed to perform their compositions to get the proximal operator associated to $f_0(\cdot)$.

\bibliographystyle{IEEEtran}
\bibliography{strings_all_ref,biblio}

\end{document}